\documentclass[10pt,journal,letterpaper,twoside,compsoc]{IEEEtran}

\usepackage{graphicx,amsmath,amssymb,verbatim,xspace}
\usepackage{color,subcaption,tabu,float,rotating,url}
\usepackage[colorlinks=true,bookmarks=false]{hyperref}
\newcommand{\COCO}{MS COCO\xspace}
\newcommand{\myparagraph}[1]{\textbf{#1}~}
\newcommand{\mysubsection}[1]{\subsection{#1}}
\newcommand{\myappendix}{the appendix\xspace}

\begin{document}
\title{Microsoft COCO: Common Objects in Context}
\author{Tsung-Yi~Lin \quad Michael~Maire \quad Serge~Belongie \quad Lubomir~Bourdev \quad Ross~Girshick \\
James~Hays \quad Pietro~Perona \quad Deva~Ramanan \quad C.~Lawrence~Zitnick \quad Piotr~Doll\'ar
\IEEEcompsocitemizethanks{
\IEEEcompsocthanksitem T.Y.~Lin and S.~Belongie are with Cornell NYC Tech and the Cornell Computer Science Department.
\IEEEcompsocthanksitem M.~Maire is with the Toyota Technological Institute at Chicago.
\IEEEcompsocthanksitem L.~Bourdev and P.~Doll\'ar are with Facebook AI Research. The majority of this work was performed while P.~Doll\'ar was with Microsoft Research.
\IEEEcompsocthanksitem R.~Girshick and C.~L.~Zitnick are with Microsoft Research, Redmond.
\IEEEcompsocthanksitem J.~Hays is with Brown University.
\IEEEcompsocthanksitem P.~Perona is with the California Institute of Technology.
\IEEEcompsocthanksitem D.~Ramanan is with the University of California at Irvine.}
}

\IEEEcompsoctitleabstractindextext{\begin{abstract}
We present a new dataset with the goal of advancing the state-of-the-art in object recognition by placing the question of object recognition in the context of the broader question of scene understanding. This is achieved by gathering images of complex everyday scenes containing common objects in their natural context. Objects are labeled using per-instance segmentations to aid in precise object localization. Our dataset contains photos of 91 objects types that would be easily recognizable by a 4 year old. With a total of 2.5 million labeled instances in 328k images, the creation of our dataset drew upon extensive crowd worker involvement via novel user interfaces for category detection, instance spotting and instance segmentation. We present a detailed statistical analysis of the dataset in comparison to PASCAL, ImageNet, and SUN. Finally, we provide baseline performance analysis for bounding box and segmentation detection results using a Deformable Parts Model.
\end{abstract}}
\maketitle

\section{Introduction}

One of the primary goals of computer vision is the understanding of visual scenes. Scene understanding involves numerous tasks including recognizing what objects are present, localizing the objects in 2D and 3D, determining the objects' and scene's attributes, characterizing relationships between objects and providing a semantic description of the scene. The current object classification and detection datasets \cite{Imagenet,PASCAL,SUN,Dollar2012PAMI} help us explore the first challenges related to scene understanding. For instance the ImageNet dataset \cite{Imagenet}, which contains an unprecedented number of images, has recently enabled breakthroughs in both object classification and detection research \cite{Hinton,GirshickDDM13,OverFeat}. The community has also created datasets containing object attributes \cite{farhadi2009describing}, scene attributes \cite{Patterson2012SunAttributes}, keypoints \cite{bourdev2009poselets}, and 3D scene information \cite{NYUDepth}. This leads us to the obvious question: what datasets will best continue our advance towards our ultimate goal of scene understanding?

\begin{figure}[!t]\centering
  \includegraphics[width=0.5\textwidth]{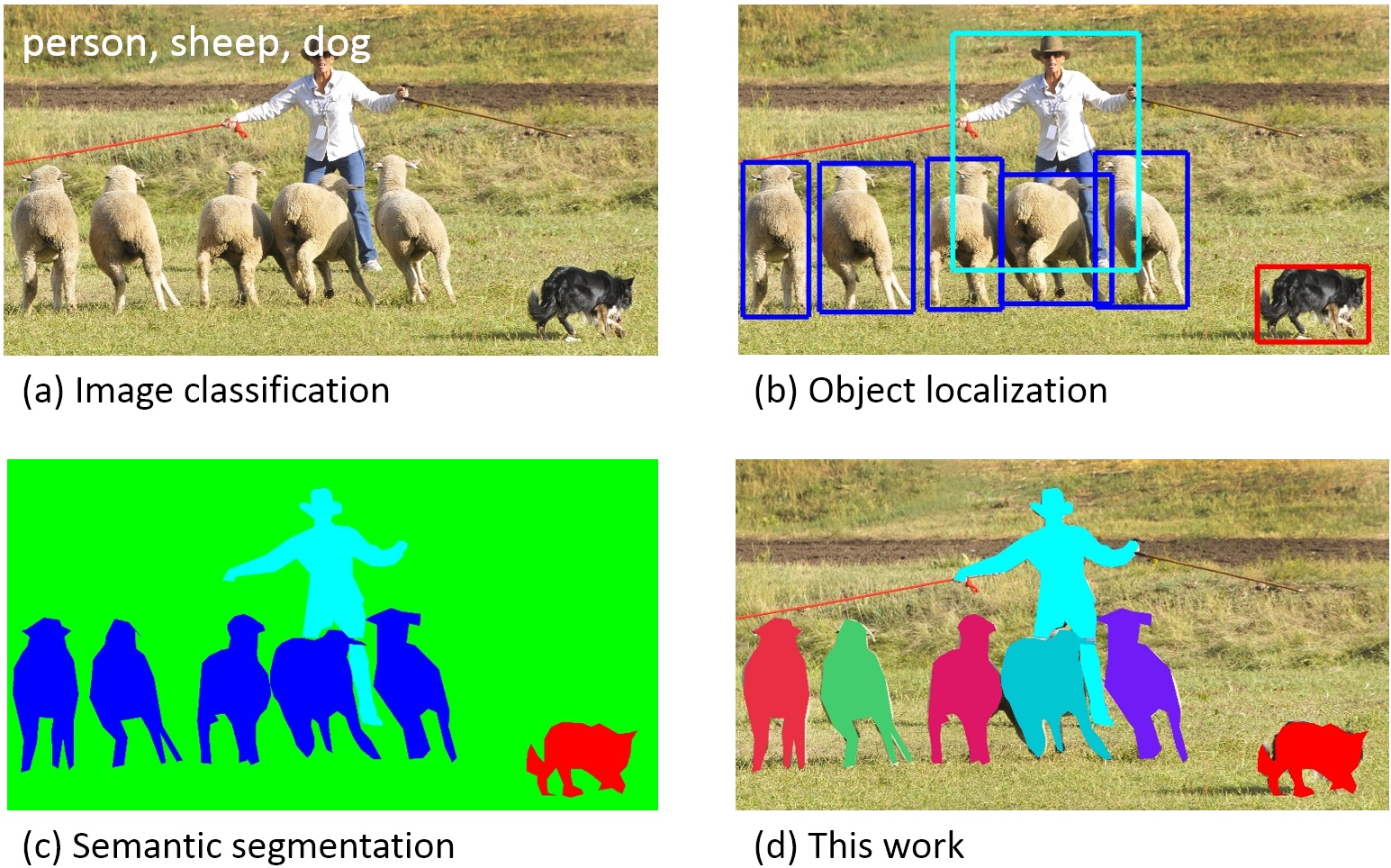}
  \caption{While previous object recognition datasets have focused on (a) image classification, (b) object bounding box localization or (c) semantic pixel-level segmentation, we focus on (d) segmenting individual object instances. We introduce a large, richly-annotated dataset comprised of images depicting complex everyday scenes of common objects in their natural context.\label{fig:teaser}}
\end{figure}

We introduce a new large-scale dataset that addresses three core research problems in scene understanding: detecting non-iconic views (or non-canonical perspectives \cite{Palmer1981}) of objects, contextual reasoning between objects and the precise 2D localization of objects. For many categories of objects, there exists an iconic view. For example, when performing a web-based image search for the object category ``bike,'' the top-ranked retrieved examples appear in profile, unobstructed near the center of a neatly composed photo. We posit that current recognition systems perform fairly well on iconic views, but struggle to recognize objects otherwise -- in the background, partially occluded, amid clutter \cite{diagnosing} -- reflecting the composition of actual everyday scenes. We verify this experimentally; when evaluated on everyday scenes, models trained on our data perform better than those trained with prior datasets. A challenge is finding natural images that contain multiple objects. The identity of many objects can only be resolved using context, due to small size or ambiguous appearance in the image. To push research in contextual reasoning, images depicting scenes \cite{SUN} rather than objects in isolation are necessary. Finally, we argue that detailed spatial understanding of object layout will be a core component of scene analysis. An object's spatial location can be defined coarsely using a bounding box \cite{PASCAL} or with a precise pixel-level segmentation \cite{brostow2009semantic,LabelMe,bell13opensurfaces}. As we demonstrate, to measure either kind of localization performance it is essential for the dataset to have every instance of every object category labeled and fully segmented. Our dataset is unique in its annotation of instance-level segmentation masks, Fig.~\ref{fig:teaser}.

To create a large-scale dataset that accomplishes these three goals we employed a novel pipeline for gathering data with extensive use of Amazon Mechanical Turk. First and most importantly, we harvested a large set of images containing contextual relationships and non-iconic object views. We accomplished this using a surprisingly simple yet effective technique that queries for pairs of objects in conjunction with images retrieved via scene-based queries \cite{ordonez2011im2text,SUN}. Next, each image was labeled as containing particular object categories using a hierarchical labeling approach \cite{Olga}. For each category found, the individual instances were labeled, verified, and finally segmented. Given the inherent ambiguity of labeling, each of these stages has numerous tradeoffs that we explored in detail.

The Microsoft Common Objects in COntext (\COCO) dataset contains 91 common object categories with 82 of them having more than 5,000 labeled instances, Fig.~\ref{fig:exampleimages}. In total the dataset has 2,500,000 labeled instances in 328,000 images. In contrast to the popular ImageNet dataset \cite{Imagenet}, COCO has fewer categories but more instances per category. This can aid in learning detailed object models capable of precise 2D localization. The dataset is also significantly larger in number of instances per category than the PASCAL VOC \cite{PASCAL} and SUN \cite{SUN} datasets. Additionally, a critical distinction between our dataset and others is the number of labeled instances per image which may aid in learning contextual information, Fig.~\ref{fig:dataanalysis}. \COCO contains considerably more object instances per image (7.7) as compared to ImageNet (3.0) and PASCAL (2.3). In contrast, the SUN dataset, which contains significant contextual information, has over 17 objects and ``stuff'' per image but considerably fewer object instances overall.

An abridged version of this work appeared in~\cite{eccv}.

\section{Related Work}

\begin{figure*}[!t]\centering
  \includegraphics[width=\textwidth]{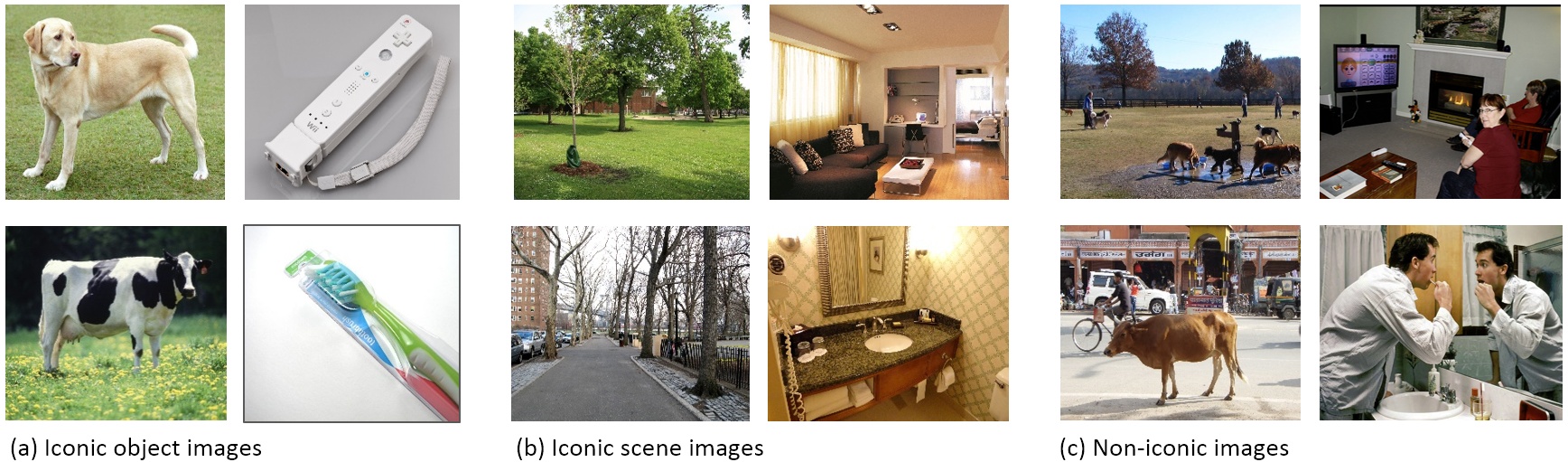}
  \caption{Example of (a) iconic object images, (b) iconic scene images, and (c) non-iconic images.\label{fig:iconic}}
\end{figure*}

Throughout the history of computer vision research datasets have played a critical role. They not only provide a means to train and evaluate algorithms, they drive research in new and more challenging directions. The creation of ground truth stereo and optical flow datasets \cite{scharstein2002taxonomy,baker2011database} helped stimulate a flood of interest in these areas. The early evolution of object recognition datasets \cite{Caltech101,Caltech256,Dalal} facilitated the direct comparison of hundreds of image recognition algorithms while simultaneously pushing the field towards more complex problems.  Recently, the ImageNet dataset \cite{Imagenet} containing millions of images has enabled breakthroughs in both object classification and detection research using a new class of deep learning algorithms \cite{Hinton,GirshickDDM13,OverFeat}.

Datasets related to object recognition can be roughly split into three groups: those that primarily address object classification, object detection and semantic scene labeling. We address each in turn.

\myparagraph{Image Classification} The task of object classification requires binary labels indicating whether objects are present in an image; see Fig.~\ref{fig:teaser}(a). Early datasets of this type comprised images containing a single object with blank backgrounds, such as the MNIST handwritten digits \cite{mnist} or COIL household objects \cite{nene1996columbia}. Caltech 101 \cite{Caltech101} and Caltech 256 \cite{Caltech256} marked the transition to more realistic object images retrieved from the internet while also increasing the number of object categories to 101 and 256, respectively. Popular datasets in the machine learning community due to the larger number of training examples, CIFAR-10 and CIFAR-100 \cite{krizhevsky2009learning} offered 10 and 100 categories from a dataset of tiny $32 \times 32$ images \cite{torralba200880}. While these datasets contained up to 60,000 images and hundreds of categories, they still only captured a small fraction of our visual world.

Recently, ImageNet \cite{Imagenet} made a striking departure from the incremental increase in dataset sizes. They proposed the creation of a dataset containing 22k categories with 500-1000 images each. Unlike previous datasets containing entry-level categories \cite{ordonezlarge}, such as ``dog'' or ``chair,'' like \cite{torralba200880}, ImageNet used the WordNet Hierarchy \cite{wordnet} to obtain both entry-level and fine-grained \cite{Birds200} categories. Currently, the ImageNet dataset contains over 14 million labeled images and has enabled significant advances in image classification \cite{Hinton,GirshickDDM13,OverFeat}.

\myparagraph{Object detection} Detecting an object entails both stating that an object belonging to a specified class is present, and localizing it in the image. The location of an object is typically represented by a bounding box, Fig.~\ref{fig:teaser}(b). Early algorithms focused on face detection \cite{hjelmaas2001face} using various ad hoc datasets. Later, more realistic and challenging face detection datasets were created \cite{LFWTech}. Another popular challenge is the detection of pedestrians for which several datasets have been created \cite{Dalal,Dollar2012PAMI}. The Caltech Pedestrian Dataset \cite{Dollar2012PAMI} contains 350,000 labeled instances with bounding boxes.

For the detection of basic object categories, a multi-year effort from 2005 to 2012 was devoted to the creation and maintenance of a series of benchmark datasets that were widely adopted. The PASCAL VOC \cite{PASCAL} datasets contained 20 object categories spread over 11,000 images. Over 27,000 object instance bounding boxes were labeled, of which almost 7,000 had detailed segmentations. Recently, a detection challenge has been created from 200 object categories using a subset of 400,000 images from ImageNet \cite{ILSVRCanalysis_ICCV2013}. An impressive 350,000 objects have been labeled using bounding boxes.

Since the detection of many objects such as sunglasses, cellphones or chairs is highly dependent on contextual information, it is important that detection datasets contain objects in their natural environments. In our dataset we strive to collect images rich in contextual information. The use of bounding boxes also limits the accuracy for which detection algorithms may be evaluated. We propose the use of fully segmented instances to enable more accurate detector evaluation.

\vspace{1mm}

\myparagraph{Semantic scene labeling} The task of labeling semantic objects in a scene requires that each pixel of an image be labeled as belonging to a category, such as sky, chair, floor, street, etc. In contrast to the detection task, individual instances of objects do not need to be segmented, Fig.~\ref{fig:teaser}(c). This enables the labeling of objects for which individual instances are hard to define, such as grass, streets, or walls. Datasets exist for both indoor \cite{NYUDepth} and outdoor \cite{shotton2009textonboost,brostow2009semantic} scenes. Some datasets also include depth information \cite{NYUDepth}. Similar to semantic scene labeling, our goal is to measure the pixel-wise accuracy of object labels. However, we also aim to distinguish between individual instances of an object, which requires a solid understanding of each object's extent.

A novel dataset that combines many of the properties of both object detection and semantic scene labeling datasets is the SUN dataset \cite{SUN} for scene understanding. SUN contains 908 scene categories from the WordNet dictionary \cite{wordnet} with segmented objects. The 3,819 object categories span those common to object detection datasets (person, chair, car) and to semantic scene labeling (wall, sky, floor). Since the dataset was collected by finding images depicting various scene types, the number of instances per object category exhibits the long tail phenomenon. That is, a few categories have a large number of instances (wall: 20,213, window: 16,080, chair: 7,971) while most have a relatively modest number of instances (boat: 349, airplane: 179, floor lamp: 276). In our dataset, we ensure that each object category has a significant number of instances, Fig.~\ref{fig:dataanalysis}.

\myparagraph{Other vision datasets} Datasets have spurred the advancement of numerous fields in computer vision. Some notable datasets include the Middlebury datasets for stereo vision \cite{scharstein2002taxonomy}, multi-view stereo \cite{seitz2006comparison} and optical flow \cite{baker2011database}. The Berkeley Segmentation Data Set (BSDS500) \cite{amfm_pami2011} has been used extensively to evaluate both segmentation and edge detection algorithms. Datasets have also been created to recognize both scene \cite{Patterson2012SunAttributes} and object attributes \cite{farhadi2009describing,lampert2009learning}. Indeed, numerous areas of vision have benefited from challenging datasets that helped catalyze progress.

\begin{figure*}[!t]\centering
  \includegraphics[width=\textwidth]{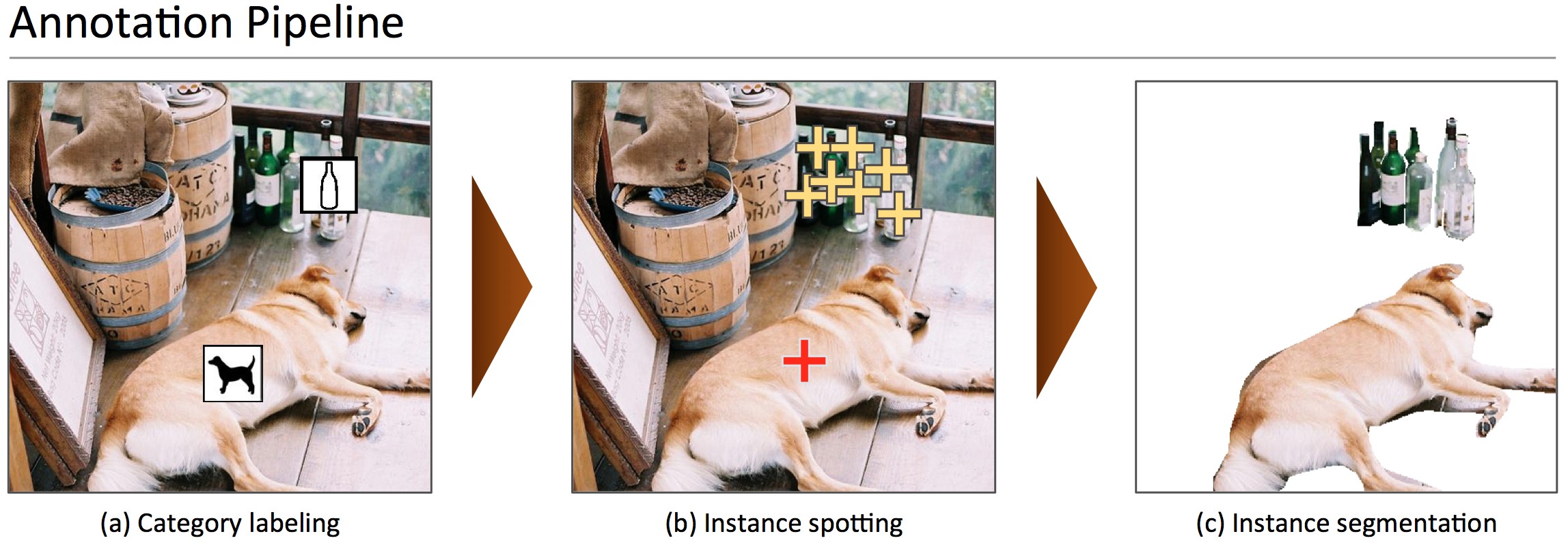}
  \caption{Our annotation pipeline is split into 3 primary tasks: (a) labeling the categories present in the image (\S\ref{sec:category-labeling}), (b) locating and marking all instances of the labeled categories (\S\ref{sec:instance-spotting}), and (c) segmenting each object instance (\S\ref{sec:instance-segmentation}).\label{fig:pipeline}}\vspace{-2mm}
\end{figure*}

\section{Image Collection}\label{sec:image_collection}

We next describe how the object categories and candidate images are selected.

\mysubsection{Common Object Categories} The selection of object categories is a non-trivial exercise. The categories must form a representative set of all categories, be relevant to practical applications and occur with high enough frequency to enable the collection of a large dataset. Other important decisions are whether to include both ``thing'' and ``stuff'' categories \cite{heitz2008learning} and whether fine-grained \cite{Birds200,Imagenet} and object-part categories should be included. ``Thing'' categories include objects for which individual instances may be easily labeled (person, chair, car) where ``stuff'' categories include materials and objects with no clear boundaries (sky, street, grass). Since we are primarily interested in precise localization of object instances, we decided to only include ``thing'' categories and not ``stuff.'' However, since ``stuff'' categories can provide significant contextual information, we believe the future labeling of ``stuff'' categories would be beneficial.

The specificity of object categories can vary significantly. For instance, a dog could be a member of the ``mammal'', ``dog'', or ``German shepherd'' categories. To enable the practical collection of a significant number of instances per category, we chose to limit our dataset to entry-level categories, i.e. category labels that are commonly used by humans when describing objects (dog, chair, person). It is also possible that some object categories may be parts of other object categories. For instance, a face may be part of a person. We anticipate the inclusion of object-part categories (face, hands, wheels) would be beneficial for many real-world applications.

\begin{figure*}[!t]\centering
  \includegraphics[width=\textwidth]{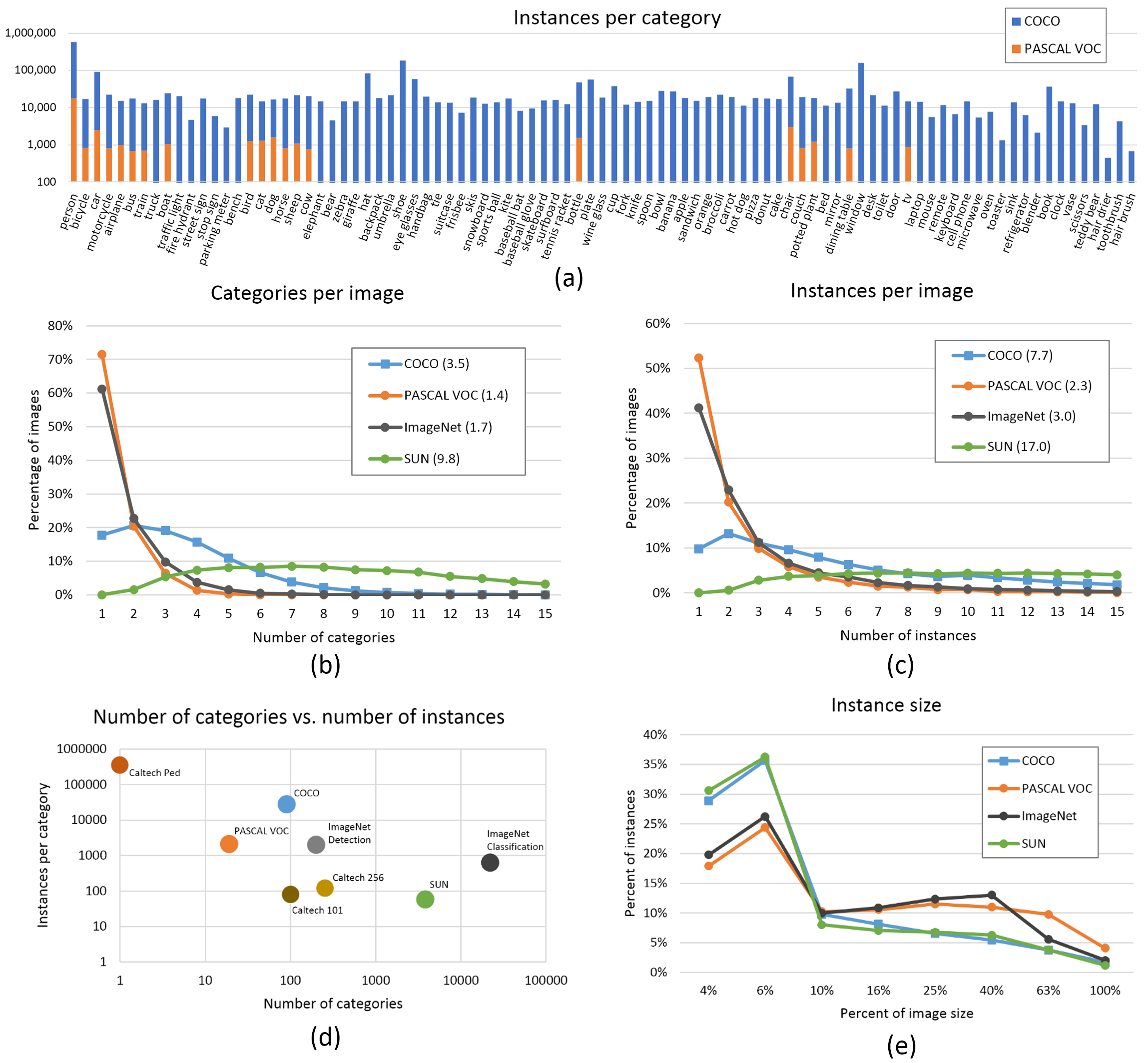}\vspace{2mm}
  \caption{(a) Number of annotated instances per category for \COCO and PASCAL VOC. (b,c) Number of annotated categories and annotated instances, respectively, per image for \COCO, ImageNet Detection, PASCAL VOC and SUN (average number of categories and instances are shown in parentheses). (d) Number of categories vs. the number of instances per category for a number of popular object recognition datasets. (e) The distribution of instance sizes for the \COCO, ImageNet Detection, PASCAL VOC and SUN datasets.\label{fig:dataanalysis}}\vspace{3mm}
\end{figure*}

We used several sources to collect entry-level object categories of ``things.'' We first compiled a list of categories by combining categories from PASCAL VOC \cite{PASCAL} and a subset of the 1200 most frequently used words that denote visually identifiable objects \cite{wordbank}. To further augment our set of candidate categories, several children ranging in ages from 4 to 8 were asked to name every object they see in indoor and outdoor environments. The final 272 candidates may be found in \myappendix. Finally, the co-authors voted on a 1 to 5 scale for each category taking into account how commonly they occur, their usefulness for practical applications, and their diversity relative to other categories. The final selection of categories attempts to pick categories with high votes, while keeping the number of categories per super-category (animals, vehicles, furniture, etc.) balanced. Categories for which obtaining a large number of instances (greater than 5,000) was difficult were also removed. To ensure backwards compatibility all categories from PASCAL VOC \cite{PASCAL} are also included. Our final list of 91 proposed categories is in Fig.~\ref{fig:dataanalysis}(a).

\mysubsection{Non-iconic Image Collection} Given the list of object categories, our next goal was to collect a set of candidate images. We may roughly group images into three types, Fig.~\ref{fig:iconic}: iconic-object images \cite{berg2009finding}, iconic-scene images \cite{SUN} and non-iconic images. Typical iconic-object images have a single large object in a canonical perspective centered in the image, Fig.~\ref{fig:iconic}(a). Iconic-scene images are shot from canonical viewpoints and commonly lack people, Fig.~\ref{fig:iconic}(b). Iconic images have the benefit that they may be easily found by directly searching for specific categories using Google or Bing image search. While iconic images generally provide high quality object instances, they can lack important contextual information and non-canonical viewpoints.

Our goal was to collect a dataset such that a majority of images are non-iconic, Fig.~\ref{fig:iconic}(c). It has been shown that datasets containing more non-iconic images are better at generalizing \cite{torralba2011unbiased}. We collected non-iconic images using two strategies. First as popularized by PASCAL VOC \cite{PASCAL}, we collected images from Flickr which tends to have fewer iconic images. Flickr contains photos uploaded by amateur photographers with searchable metadata and keywords. Second, we did not search for object categories in isolation. A search for ``dog'' will tend to return iconic images of large, centered dogs. However, if we searched for pairwise combinations of object categories, such as ``dog + car'' we found many more non-iconic images. Surprisingly, these images typically do not just contain the two categories specified in the search, but numerous other categories as well. To further supplement our dataset we also searched for scene/object category pairs, see \myappendix. We downloaded at most 5 photos taken by a single photographer within a short time window. In the rare cases in which enough images could not be found, we searched for single categories and performed an explicit filtering stage to remove iconic images. The result is a collection of 328,000 images with rich contextual relationships between objects as shown in Figs.~\ref{fig:iconic}(c) and \ref{fig:exampleimages}.

\section{Image Annotation}

\begin{figure*}[!t]\centering
  \begin{subfigure}[b]{0.47\textwidth}
  \includegraphics[width=\textwidth]{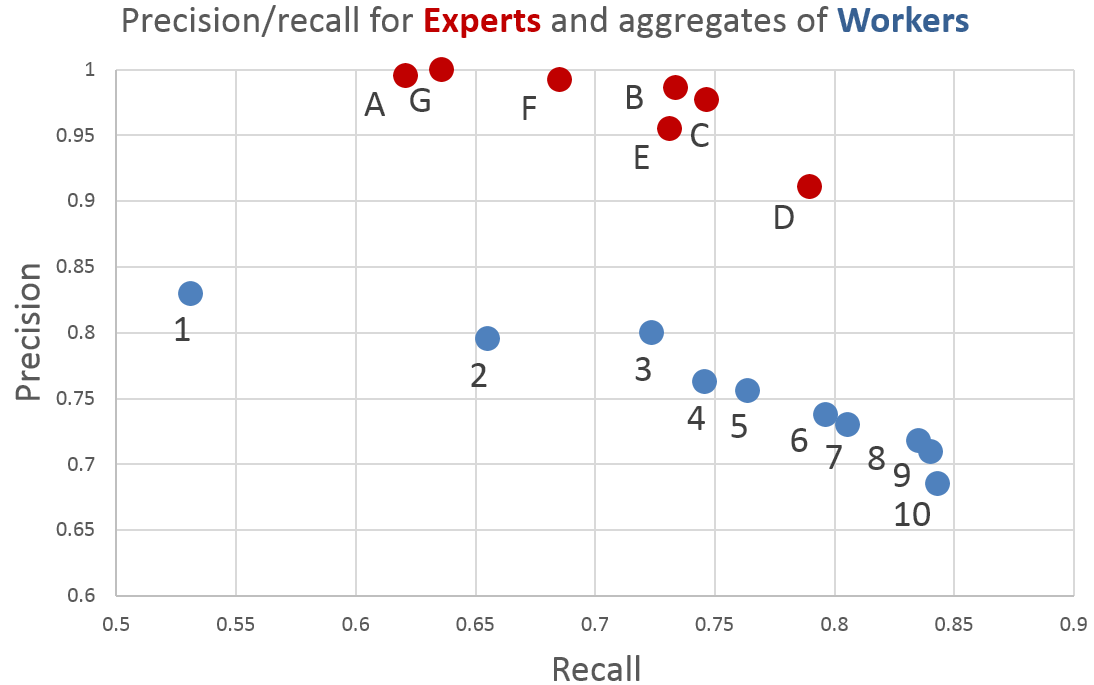}\vspace{-3mm}
  \label{fig:recall}\caption{}
  \end{subfigure}\quad\
  \begin{subfigure}[b]{0.50\textwidth}
  \includegraphics[width=\textwidth]{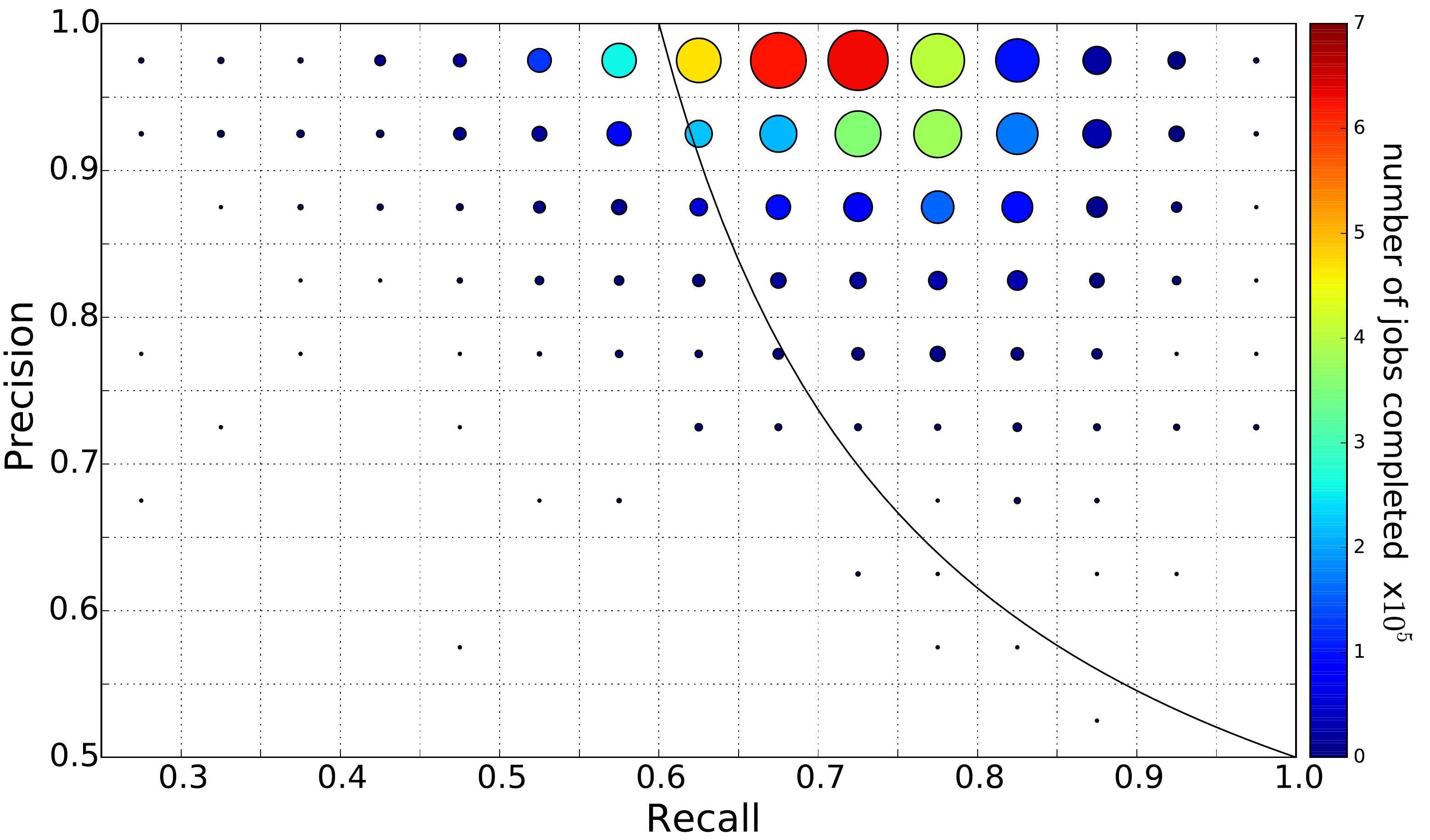}\vspace{-3mm}
  \label{fig:all_worker}\caption{}
  \end{subfigure}
  \caption{Worker precision and recall for the category labeling task. (a) The union of multiple AMT workers (blue) has better recall than any expert (red). Ground truth was computed using majority vote of the experts. (b) Shows the number of workers (circle size) and average number of jobs per worker (circle color) for each precision/recall range. Most workers have high precision; such workers generally also complete more jobs. For this plot ground truth for each worker is the \emph{union} of responses from all other AMT workers. See \S\ref{sec:annotation-performance} for details.\label{fig:workers}}
\end{figure*}

We next describe how we annotated our image collection. Due to our desire to label over 2.5 million object instances, the design of a cost efficient yet high quality annotation pipeline was critical. The annotation pipeline is outlined in Fig.~\ref{fig:pipeline}. For all crowdsourcing tasks we used workers on Amazon's Mechanical Turk (AMT). Our user interfaces are described in detail in \myappendix. Note that, since the original version of this work \cite{eccv}, we have taken a number of steps to further improve the quality of the annotations. In particular, we have increased the number of annotators for the category labeling and instance spotting stages to eight. We also added a stage to verify the instance segmentations.

\mysubsection{Category Labeling}\label{sec:category-labeling} The first task in annotating our dataset is determining which object categories are present in each image, Fig.~\ref{fig:pipeline}(a). Since we have 91 categories and a large number of images, asking workers to answer 91 binary classification questions per image would be prohibitively expensive. Instead, we used a hierarchical approach \cite{Olga}.

We group the object categories into 11 super-categories (see \myappendix). For a given image, a worker was presented with each group of categories in turn and asked to indicate whether any instances exist for that super-category. This greatly reduces the time needed to classify the various categories. For example, a worker may easily determine no animals are present in the image without having to specifically look for cats, dogs, etc. If a worker determines instances from the super-category (animal) are present, for each subordinate category (dog, cat, etc.) present, the worker must drag the category's icon onto the image over one instance of the category. The placement of these icons is critical for the following stage. We emphasize that only a single instance of each category needs to be annotated in this stage. To ensure high recall, 8 workers were asked to label each image. A category is considered present if any worker indicated the category; false positives are handled in subsequent stages. A detailed analysis of performance is presented in \S\ref{sec:annotation-performance}. This stage took $\sim$20k worker hours to complete.

\mysubsection{Instance Spotting}\label{sec:instance-spotting} In the next stage all instances of the object categories in an image were labeled, Fig.~\ref{fig:pipeline}(b). In the previous stage each worker labeled one instance of a category, but multiple object instances may exist. Therefore, for each image, a worker was asked to place a cross on top of each instance of a specific category found in the previous stage. To boost recall, the location of the instance found by a worker in the previous stage was shown to the current worker. Such priming helped workers quickly find an initial instance upon first seeing the image. The workers could also use a magnifying glass to find small instances. Each worker was asked to label at most 10 instances of a given category per image. Each image was labeled by 8 workers for a total of $\sim$10k worker hours.

\mysubsection{Instance Segmentation}\label{sec:instance-segmentation} Our final stage is the laborious task of segmenting each object instance, Fig.~\ref{fig:pipeline}(c). For this stage we modified the excellent user interface developed by Bell et al.~\cite{bell13opensurfaces} for image segmentation. Our interface asks the worker to segment an object instance specified by a worker in the previous stage. If other instances have already been segmented in the image, those segmentations are shown to the worker. A worker may also indicate there are no object instances of the given category in the image (implying a false positive label from the previous stage) or that all object instances are already segmented.

\begin{figure*}[t]\centering
  \includegraphics[width=\textwidth]{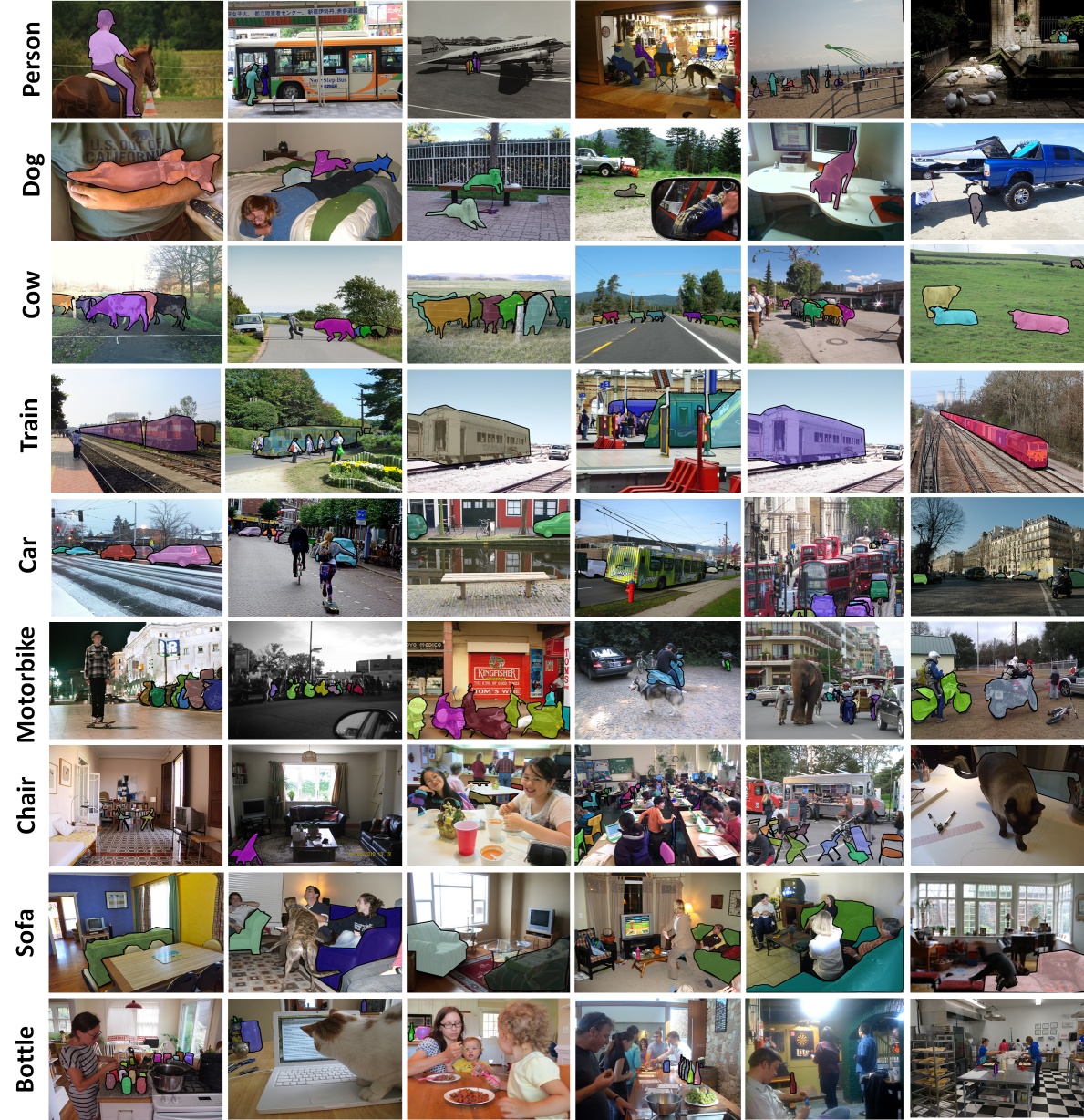}
  \caption{Samples of annotated images in the \COCO dataset.\label{fig:exampleimages}}
\end{figure*}

Segmenting 2,500,000 object instances is an extremely time consuming task requiring over 22 worker hours per 1,000 segmentations. To minimize cost we only had a single worker segment each instance. However, when first completing the task, most workers produced only coarse instance outlines. As a consequence, we required all workers to complete a training task for each object category. The training task required workers to segment an object instance. Workers could not complete the task until their segmentation adequately matched the ground truth. The use of a training task vastly improved the quality of the workers (approximately 1 in 3 workers passed the training stage) and resulting segmentations. Example segmentations may be viewed in Fig.~\ref{fig:exampleimages}.

While the training task filtered out most bad workers, we also performed an explicit verification step on each segmented instance to ensure good quality. Multiple workers (3 to 5) were asked to judge each segmentation and indicate whether it matched the instance well or not. Segmentations of insufficient quality were discarded and the corresponding instances added back to the pool of unsegmented objects. Finally, some approved workers consistently produced poor segmentations; all work obtained from such workers was discarded.

For images containing 10 object instances or fewer of a given category, every instance was individually segmented (note that in some images up to 15 instances were segmented). Occasionally the number of instances is drastically higher; for example, consider a dense crowd of people or a truckload of bananas. In such cases, many instances of the same category may be tightly grouped together and distinguishing individual instances is difficult. After 10-15 instances of a category were segmented in an image, the remaining instances were marked as ``crowds'' using a single (possibly multi-part) segment. For the purpose of evaluation, areas marked as crowds will be ignored and not affect a detector's score. Details are given in \myappendix.

\mysubsection{Annotation Performance Analysis}\label{sec:annotation-performance} We analyzed crowd worker quality on the category labeling task by comparing to dedicated expert workers, see Fig.~\ref{fig:workers}(a). We compared precision and recall of seven expert workers (co-authors of the paper) with the results obtained by taking the union of one to ten AMT workers. Ground truth was computed using majority vote of the experts. For this task recall is of primary importance as false positives could be removed in later stages. Fig.~\ref{fig:workers}(a) shows that the union of 8 AMT workers, the same number as was used to collect our labels, achieved greater recall than any of the expert workers. Note that worker recall saturates at around 9-10 AMT workers.

Object category presence is often ambiguous. Indeed as Fig.~\ref{fig:workers}(a) indicates, even dedicated experts often disagree on object presence, e.g.~due to inherent ambiguity in the image or disagreement about category definitions. For any unambiguous examples having a probability of over 50\% of being annotated, the probability all 8 annotators missing such a case is at most $.5^8 \approx .004$. Additionally, by observing how recall increased as we added annotators, we estimate that in practice over 99\% of all object categories not later rejected as false positives are detected given 8 annotators. Note that a similar analysis may be done for instance spotting in which 8 annotators were also used.

Finally, Fig.~\ref{fig:workers}(b) re-examines precision and recall of AMT workers on category labeling on a much larger set of images. The number of workers (circle size) and average number of jobs per worker (circle color) is shown for each precision/recall range. Unlike in Fig.~\ref{fig:workers}(a), we used a leave-one-out evaluation procedure where a category was considered present if \emph{any} of the remaining workers named the category. Therefore, overall worker precision is substantially higher. Workers who completed the most jobs also have the highest precision; all jobs from workers below the black line were rejected.

\mysubsection{Caption Annotation}

We added five written caption descriptions to each image in \COCO. A full description of the caption statistics and how they were gathered will be provided shortly in a separate publication.

\section{Dataset Statistics}

Next, we analyze the properties of the Microsoft Common Objects in COntext (\COCO) dataset in comparison to several other popular datasets. These include ImageNet \cite{Imagenet}, PASCAL VOC 2012 \cite{PASCAL}, and SUN \cite{SUN}. Each of these datasets varies significantly in size, list of labeled categories and types of images. ImageNet was created to capture a large number of object categories, many of which are fine-grained. SUN focuses on labeling scene types and the objects that commonly occur in them. Finally, PASCAL VOC's primary application is object detection in natural images. \COCO is designed for the detection and segmentation of objects occurring in their natural context.

The number of instances per category for all 91 categories is shown in Fig.~\ref{fig:dataanalysis}(a). A summary of the datasets showing the number of object categories and the number of instances per category is shown in Fig.~\ref{fig:dataanalysis}(d). While \COCO has fewer categories than ImageNet and SUN, it has more instances per category which we hypothesize will be useful for learning complex models capable of precise localization.  In comparison to PASCAL VOC, \COCO has both more categories and instances.

An important property of our dataset is we strive to find non-iconic images containing objects in their natural context. The amount of contextual information present in an image can be estimated by examining the average number of object categories and instances per image, Fig.~\ref{fig:dataanalysis}(b, c). For ImageNet we plot the object detection validation set, since the training data only has a single object labeled. On average our dataset contains 3.5 categories and 7.7 instances per image. In comparison ImageNet and PASCAL VOC both have less than 2 categories and 3 instances per image on average. Another interesting observation is only $10\%$ of the images in \COCO have only one category per image, in comparison, over $60\%$ of images contain a single object category in ImageNet and PASCAL VOC.  As expected, the SUN dataset has the most contextual information since it is scene-based and uses an unrestricted set of categories.

Finally, we analyze the average size of objects in the datasets. Generally smaller objects are harder to recognize and require more contextual reasoning to recognize. As shown in Fig.~\ref{fig:dataanalysis}(e), the average sizes of objects is smaller for both \COCO and SUN.

\section{Dataset Splits}

To accommodate a faster release schedule, we split the \COCO dataset into two roughly equal parts. The first half of the dataset was released in 2014, the second half will be released in 2015. The 2014 release contains 82,783 training, 40,504 validation, and 40,775 testing images (approximately $\frac{1}{2}$ train, $\frac{1}{4}$ val, and $\frac{1}{4}$ test). There are nearly 270k segmented people and a total of 886k segmented object instances in the 2014 train+val data alone. The cumulative 2015 release will contain a total of 165,482 train, 81,208 val, and 81,434 test images. We took care to minimize the chance of near-duplicate images existing across splits by explicitly removing near duplicates (detected with \cite{douze2009evaluation}) and grouping images by photographer and date taken.

Following established protocol, annotations for train and validation data will be released, but not for test. We are currently finalizing the evaluation server for automatic evaluation on the test set. A full discussion of evaluation metrics will be added once the evaluation server is complete.

Note that we have limited the 2014 release to a subset of 80 categories. We did not collect segmentations for the following 11 categories: hat, shoe, eyeglasses (too many instances), mirror, window, door, street sign (ambiguous and difficult to label), plate, desk (due to confusion with bowl and dining table, respectively) and blender, hair brush (too few instances). We may add segmentations for some of these categories in the cumulative 2015 release.

\section{Algorithmic Analysis}

\begin{table*}
{\tiny\resizebox{\textwidth}{!}{\tabcolsep=0.05cm\begin{tabu}[ht]{@{}l*{21}{c}@{} }
  \rowfont{\footnotesize} &plane &bike &bird &boat &bottle &bus &car &cat &chair &cow &table &dog &horse &moto &person &plant &sheep &sofa &train &tv &avg.\\\hline
  \tabularnewline\rowfont{\footnotesize} DPMv5-P  & {\bf 45.6} & 49.0 & 11.0 & {\bf 11.6} & {\bf 27.2} & 50.5 & {\bf 43.1} & {\bf 23.6} & {\bf 17.2} & 23.2 & {\bf 10.7} & {\bf 20.5} & 42.5 & {\bf 44.5} & {\bf 41.3} & {\bf 8.7} & {\bf 29.0} & {\bf 18.7 } & {\bf 40.0} & 34.5 & {\bf 29.6} \\
  \tabularnewline\rowfont{\footnotesize} DPMv5-C  & 43.7 & {\bf 50.1} & {\bf 11.8} & 2.4 & 21.4 & {\bf 60.1} & 35.6 & 16.0 & 11.4 & {\bf 24.8} & 5.3 & 9.4 & {\bf 44.5} & 41.0 & 35.8 & 6.3 & 28.3 & 13.3 & 38.8 & {\bf 36.2} & 26.8  \\\hline
  \tabularnewline\rowfont{\footnotesize} DPMv5-P  & 35.1 & 17.9 & 3.7 & 2.3 & {\bf 7} & 45.4 & {\bf 18.3} & 8.6 & {\bf 6.3} & 17 & 4.8 & {\bf 5.8} & 35.3 & 25.4 & {\bf 17.5} & 4.1 & {\bf 14.5} & 9.6 & 31.7 & 27.9 & 16.9\\
  \tabularnewline\rowfont{\footnotesize} DPMv5-C  & {\bf 36.9} & {\bf 20.2} & {\bf 5.7} & {\bf 3.5} & 6.6 & {\bf 50.3} & 16.1 & {\bf 12.8} & 4.5 & {\bf 19.0} & {\bf 9.6} & 4.0 & {\bf 38.2} & {\bf 29.9} & 15.9 & {\bf 6.7} & 13.8 & {\bf 10.4} & {\bf 39.2} & {\bf 37.9} & {\bf 19.1}\\
\tabularnewline\end{tabu}}}
\caption{\textbf{Top}: Detection performance evaluated on \textbf{PASCAL VOC 2012}. DPMv5-P is the performance reported by Girshick et al.~in VOC release 5. DPMv5-C uses the same implementation, but is trained with \COCO. \textbf{Bottom}: Performance evaluated on \textbf{\COCO} for DPM models trained with PASCAL VOC 2012 (DPMv5-P) and \COCO (DPMv5-C). For DPMv5-C we used 5000 positive and 10000 negative training examples. While \COCO is considerably more challenging than PASCAL, use of more training data coupled with more sophisticated approaches \cite{Hinton,GirshickDDM13,OverFeat} should improve performance substantially.\label{tab:ap_scores}}
\end{table*}

\myparagraph{Bounding-box detection} For the following experiments we take a subset of 55,000 images from our dataset\footnote{These preliminary experiments were performed before our final split of the dataset intro train, val, and test. Baselines on the actual test set will be added once the evaluation server is complete.} and obtain tight-fitting bounding boxes from the annotated segmentation masks. We evaluate models tested on both \COCO and PASCAL, see Table \ref{tab:ap_scores}. We evaluate two different models. \textbf{DPMv5-P}: the latest implementation of~\cite{felzenszwalb2010object} (release 5 \cite{voc-release5}) trained on PASCAL VOC 2012. \textbf{DPMv5-C}: the same implementation trained on COCO (5000 positive and 10000 negative images).  We use the default parameter settings for training COCO models.

If we compare the average performance of DPMv5-P on PASCAL VOC and \COCO, we find that average performance on \COCO drops by nearly a {\em factor of 2}, suggesting that \COCO does include more difficult (non-iconic) images of objects that are partially occluded, amid clutter, etc. We notice a similar drop in performance for the model trained on \COCO (DPMv5-C).

The effect on detection performance of training on PASCAL VOC or \COCO may be analyzed by comparing DPMv5-P and DPMv5-C. They use the same implementation with different sources of training data. Table \ref{tab:ap_scores} shows DPMv5-C still outperforms DPMv5-P in 6 out of 20 categories when testing on PASCAL VOC. In some categories (e.g., dog, cat, people), models trained on \COCO perform worse, while on others (e.g., bus, tv, horse), models trained on our data are better.

Consistent with past observations~\cite{zhu2012we}, we find that including difficult (non-iconic) images during training may not always help. Such examples may act as noise and pollute the learned model if the model is not rich enough to capture such appearance variability. Our dataset allows for the exploration of such issues.

Torralba and Efros \cite{torralba2011unbiased} proposed a metric to measure cross-dataset generalization which computes the `performance drop' for models that train on one dataset and test on another. The performance difference of the DPMv5-P models across the two datasets is 12.7 AP while the DPMv5-C models only have 7.7 AP difference. Moreover, overall performance is much lower on \COCO. These observations support two hypotheses: 1) \COCO is significantly more difficult than PASCAL VOC and 2) models trained on \COCO can generalize better to easier datasets such as PASCAL VOC given more training data. To gain insight into the differences between the datasets, see \myappendix for visualizations of person and chair examples from the two datasets.

\begin{figure}[!t]\centering
  \includegraphics[width=\columnwidth]{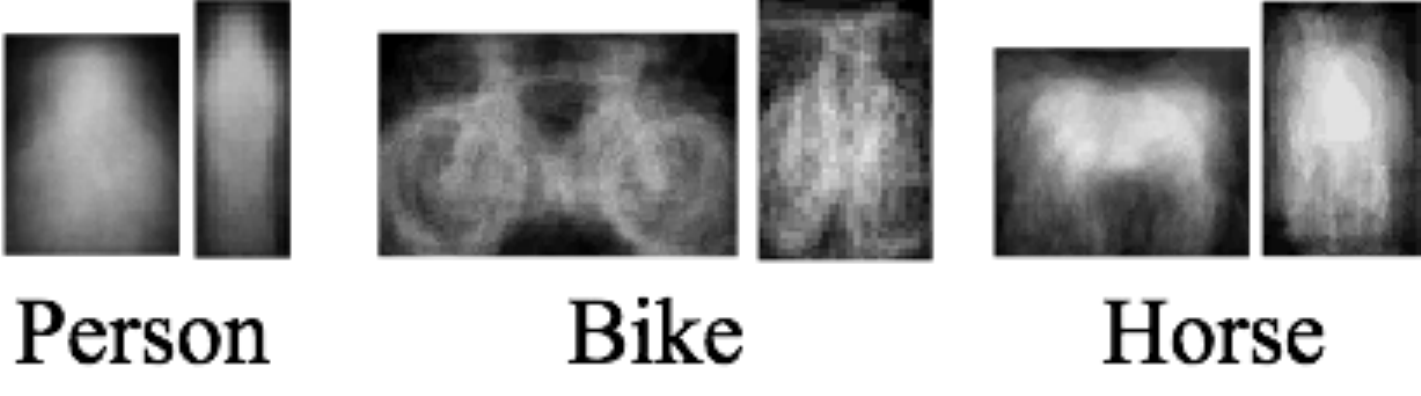}
  \caption{We visualize our mixture-specific shape masks. We paste thresholded shape masks on each candidate detection to generate candidate segments.\label{fig:masks}}
\end{figure}

\myparagraph{Generating segmentations from detections} We now describe a simple method for generating object bounding boxes and segmentation masks, following prior work that produces segmentations from object detections \cite{brox2011object,yang2012layered,ramanan2007using,dai2012learning}. We learn aspect-specific pixel-level segmentation masks for different categories. These are readily learned by averaging together segmentation masks from aligned training instances. We learn different masks corresponding to the different mixtures in our DPM detector. Sample masks are visualized in Fig.~\ref{fig:masks}.

\begin{figure}[!t]\centering
  \includegraphics[width=\columnwidth]{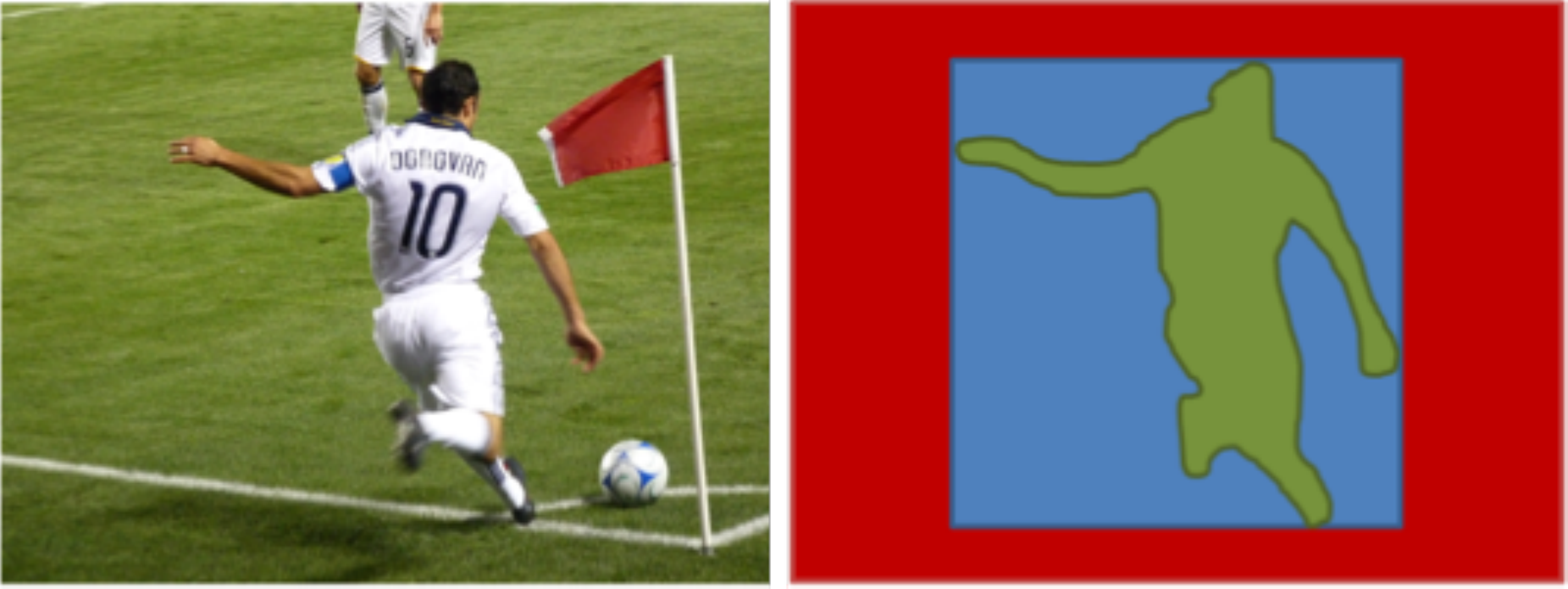}
  \caption{Evaluating instance detections with segmentation masks versus bounding boxes. Bounding boxes are a particularly crude approximation for articulated objects; in this case, the majority of the pixels in the ({\bf blue}) tight-fitting bounding-box do not lie on the object. Our ({\bf green}) instance-level segmentation masks allows for a more accurate measure of object detection and localization.\label{fig:segment}}\vspace{-2mm}
\end{figure}

\begin{figure*}[!t]\centering
  \begin{minipage}[t]{0.03\textwidth}
    \vspace{4.5\linewidth}
    \hfill\begin{sideways}\scriptsize{Predicted\hspace{.7cm}Ground truth}\end{sideways}\hfill
  \end{minipage}
  \begin{minipage}[t]{0.17\textwidth}\vspace{0pt}
    \setlength\fboxsep{0pt}
    \fbox{\includegraphics[width=\linewidth]{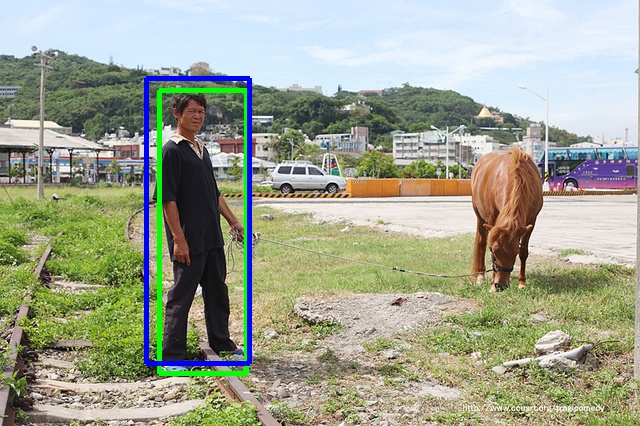}} \\
    \fbox{\includegraphics[width=\linewidth]{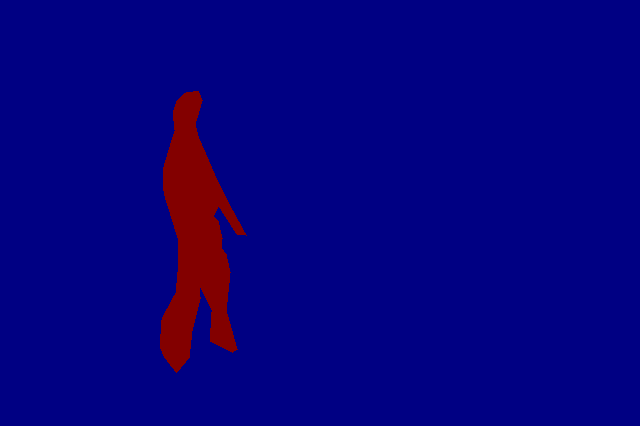}} \\
    \fbox{\includegraphics[width=\linewidth]{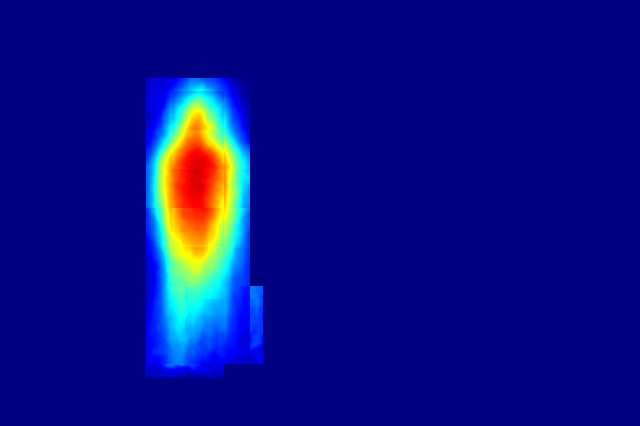}}
  \end{minipage}\hfill
  \begin{minipage}[t]{0.20\textwidth}\vspace{0pt}
    \includegraphics[width=\linewidth, clip=true, trim=2.6in 2.60in 2.6in 2.60in]{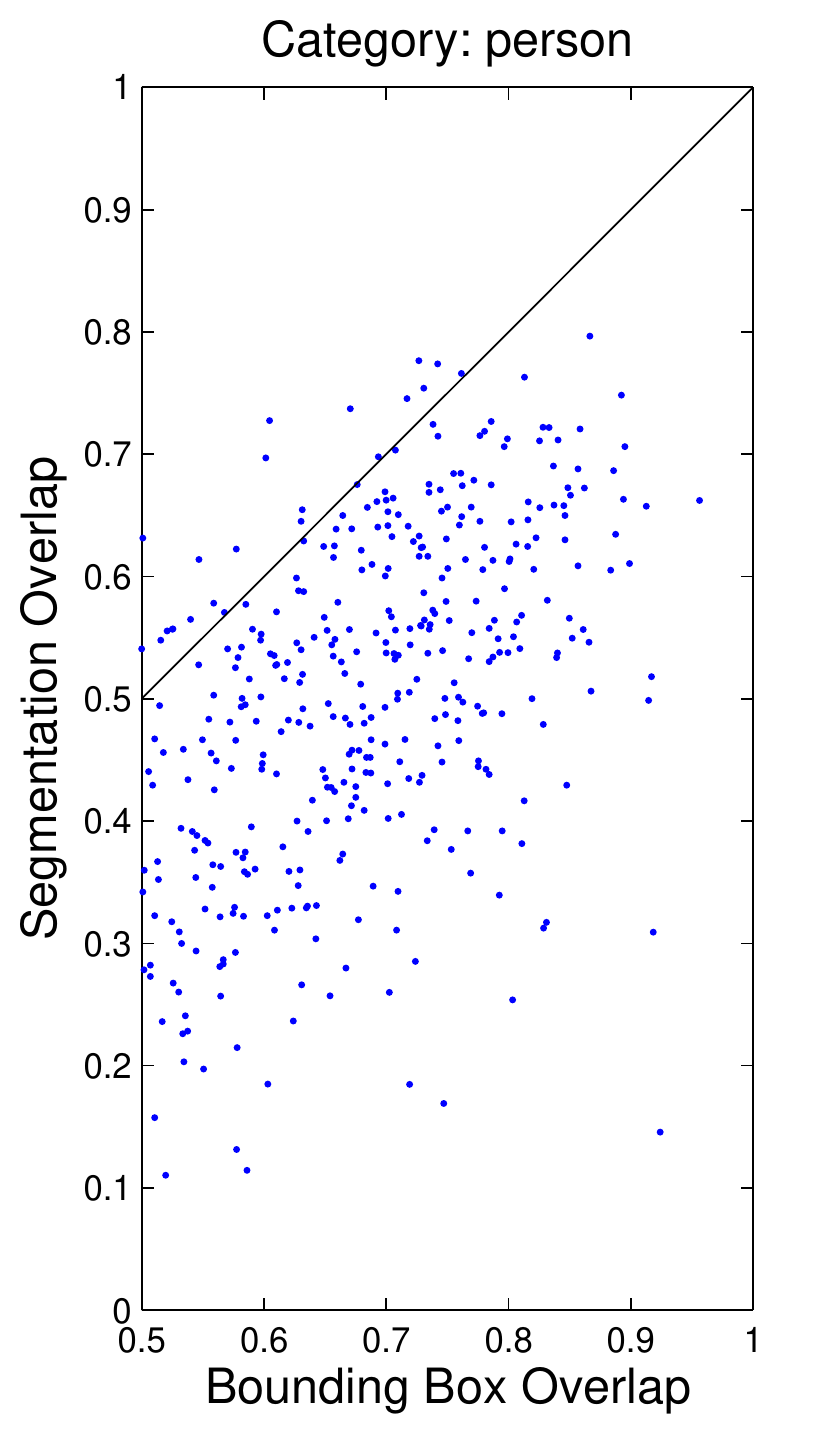}
  \end{minipage}
  \begin{minipage}[t]{0.55\textwidth}\vspace{0pt}
    \includegraphics[width=\linewidth, clip=true, trim=0.5in 2.60in 1.0in 2.60in]{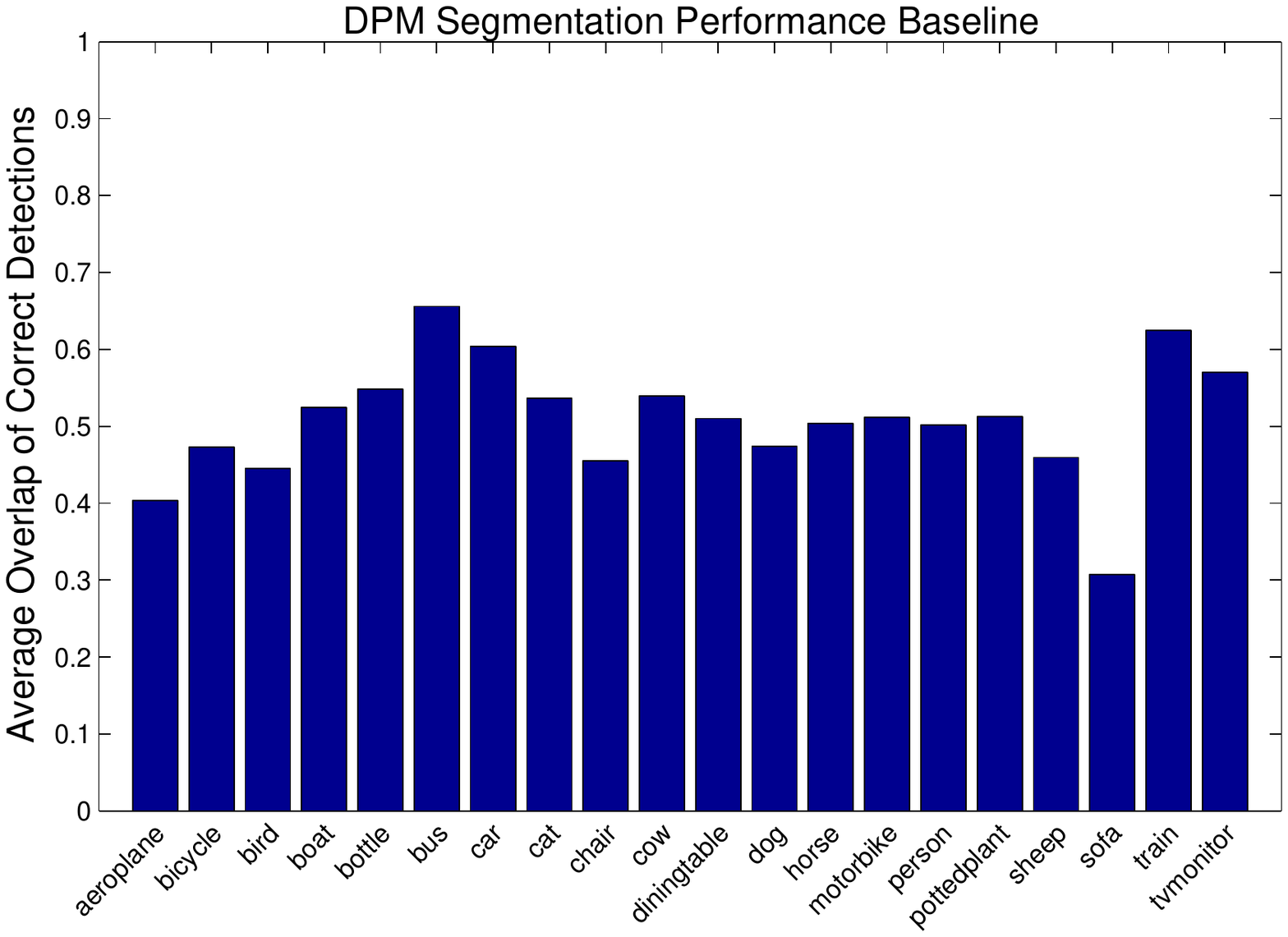}
  \end{minipage}
  \vspace{-0.08\linewidth}
  \caption{A predicted segmentation might not recover object detail even though detection and ground truth bounding boxes overlap well (left). Sampling from the person category illustrates that predicting segmentations from top-down projection of DPM part masks is difficult even for correct detections (center). Average segmentation overlap measured on \COCO for the 20 PASCAL VOC categories demonstrates the difficulty of the problem (right).\label{fig:seg_eval}}
\end{figure*}

\myparagraph{Detection evaluated by segmentation} Segmentation is a challenging task even assuming a detector reports correct results as it requires fine localization of object part boundaries. To decouple segmentation evaluation from detection correctness, we benchmark segmentation quality using only correct detections. Specifically, given that the detector reports a correct bounding box, how well does the predicted segmentation of that object match the ground truth segmentation? As criterion for correct detection, we impose the standard requirement that intersection over union between predicted and ground truth boxes is at least $0.5$. We then measure the intersection over union of the predicted and ground truth segmentation masks, see Fig.~\ref{fig:segment}. To establish a baseline for our dataset, we project learned DPM part masks onto the image to create segmentation masks. Fig.~\ref{fig:seg_eval} shows results of this segmentation baseline for the DPM learned on the 20 PASCAL categories and tested on our dataset.

\section{Discussion}

We introduced a new dataset for detecting and segmenting objects found in everyday life in their natural environments. Utilizing over 70,000 worker hours, a vast collection of object instances was gathered, annotated and organized to drive the advancement of object detection and segmentation algorithms. Emphasis was placed on finding non-iconic images of objects in natural environments and varied viewpoints. Dataset statistics indicate the images contain rich contextual information with many objects present per image.

There are several promising directions for future annotations on our dataset. We currently only label ``things'', but labeling ``stuff'' may also provide significant contextual information that may be useful for detection. Many object detection algorithms benefit from additional annotations, such as the amount an instance is occluded \cite{Dollar2012PAMI} or the location of keypoints on the object \cite{bourdev2009poselets}. Finally, our dataset could provide a good benchmark for other types of labels, including scene types \cite{SUN}, attributes \cite{Patterson2012SunAttributes,farhadi2009describing} and full sentence written descriptions \cite{rashtchian2010collecting}. We are actively exploring adding various such annotations.

To download and learn more about \COCO please see the project website\footnote{\url{http://mscoco.org/}}. \COCO will evolve and grow over time; up to date information is available online.

{\myparagraph{Acknowledgments} Funding for all crowd worker tasks was provided by Microsoft. P.P.~and D.R.~were supported by ONR MURI Grant N00014-10-1-0933. We would like to thank all members of the community who provided valuable feedback throughout the process of defining and collecting the dataset.}

\section*{Appendix Overview}

In the appendix, we provide detailed descriptions of the AMT user interfaces and the full list of 272 candidate categories (from which our final 91 were selected) and 40 scene categories (used for scene-object queries).

\section*{Appendix I: User Interfaces}

We describe and visualize our user interfaces for collecting non-iconic images, category labeling, instance spotting, instance segmentation, segmentation verification and finally crowd labeling.

\begin{figure}\centering
  \includegraphics[width=0.5\textwidth]{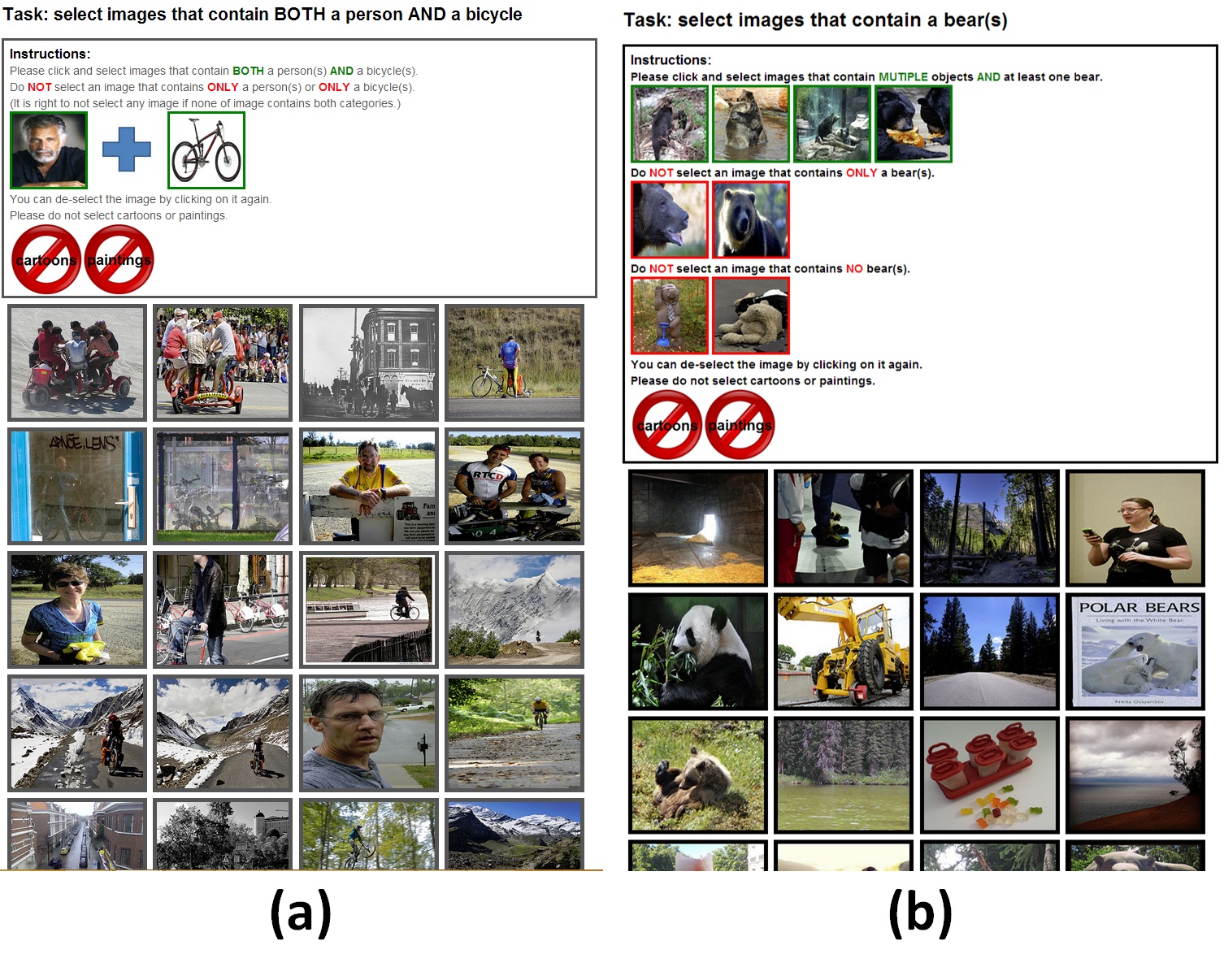}
  \caption{User interfaces for non-iconic image collection. (a) Interface for selecting non-iconic images containing pairs of objects. (b) Interface for selecting non-iconic images for categories that rarely co-occurred with others.\label{fig:ui_collection}}
\end{figure}

\myparagraph{Non-iconic Image Collection} Flickr provides a rich image collection associated with text captions. However, captions might be inaccurate and images may be iconic. To construct a high-quality set of non-iconic images, we first collected candidate images by searching for pairs of object categories, or pairs of object and scene categories. We then created an AMT filtering task that allowed users to remove invalid or iconic images from a grid of 128 candidates, Fig.~\ref{fig:ui_collection}. We found the choice of instructions to be crucial, and so provided users with examples of iconic and non-iconic images. Some categories rarely co-occurred with others. In such cases, we collected candidates using only the object category as the search term, but apply a similar filtering step, Fig.~\ref{fig:ui_collection}(b).

\begin{figure*}\centering
  \includegraphics[width=1\textwidth]{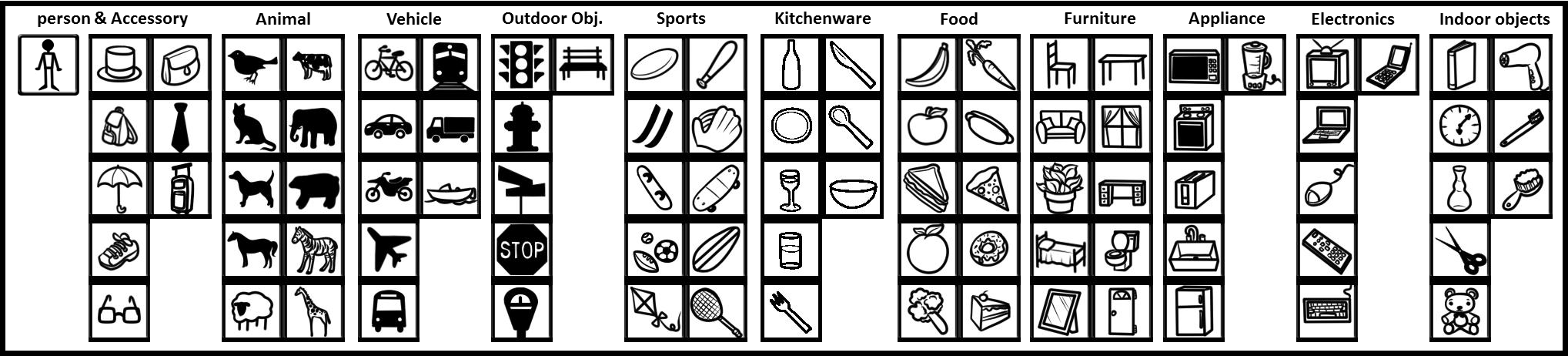}
  \caption{Icons of 91 categories in the \COCO dataset grouped by 11 super-categories. We use these icons in our annotation pipeline to help workers quickly reference the indicated object category.\label{fig:icons}}
\end{figure*}

\begin{figure*}\centering
  \includegraphics[width=1\textwidth]{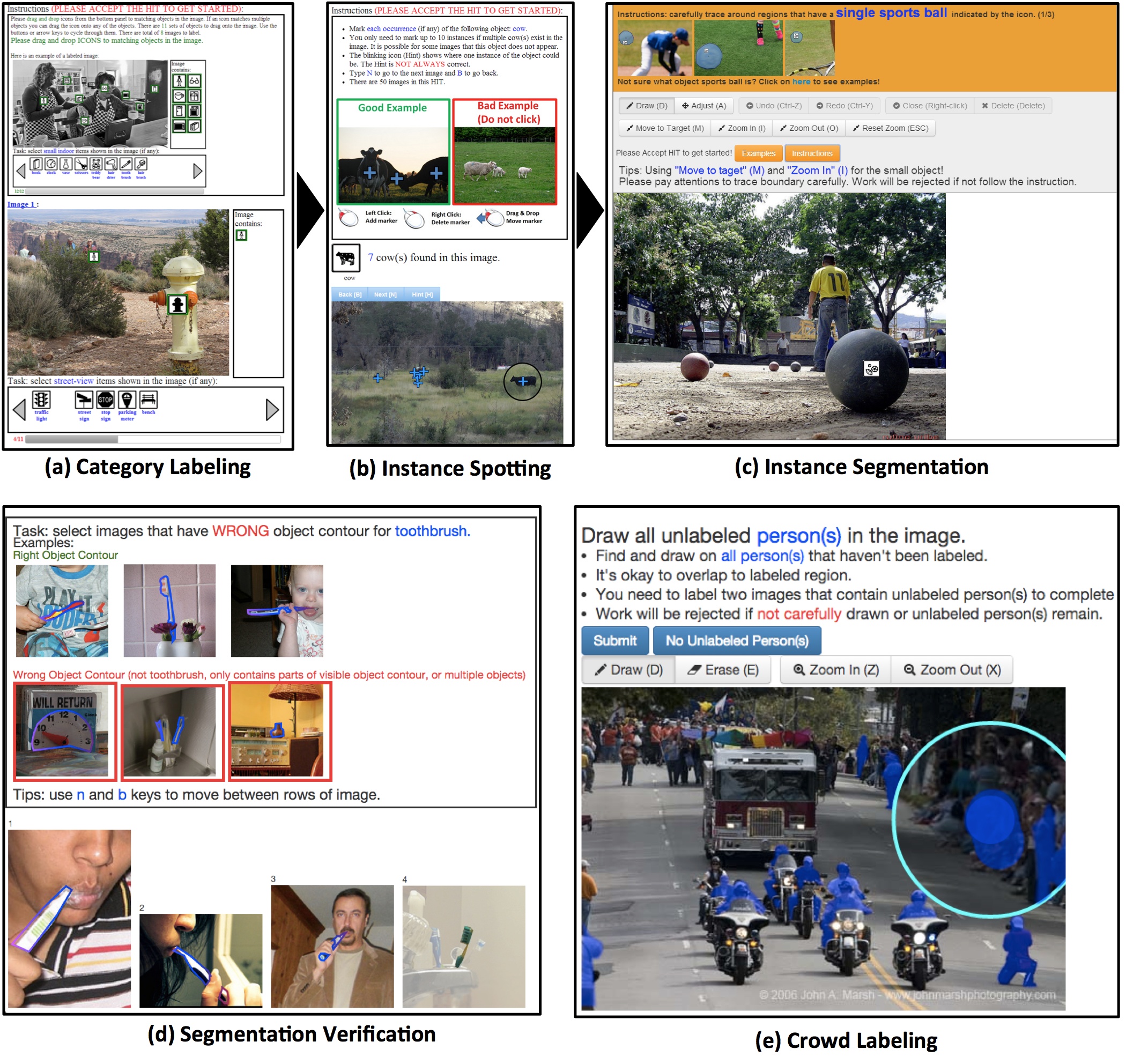}
  \caption{ User interfaces for collecting instance annotations, see text for details.\label{fig:ui_pipeline}}
\end{figure*}

\myparagraph{Category Labeling} Fig.~\ref{fig:ui_pipeline}(a) shows our interface for category labeling. We designed the labeling task to encourage workers to annotate all categories present in the image. Workers annotate categories by dragging and dropping icons from the bottom category panel onto a corresponding object instance. Only a single instance of each object category needs to be annotated in the image. We group icons by the super-categories from Fig.~\ref{fig:icons}, allowing workers to quickly skip categories that are unlikely to be present.

\myparagraph{Instance Spotting} Fig.~\ref{fig:ui_pipeline}(b) depicts our interface for labeling all instances of a given category. The interface is initialized with a blinking icon specifying a single instance obtained from the previous category-labeling stage. Workers are then asked to spot and click on up to 10 total instances of the given category, placing a single cross anywhere within the region of each instance. In order to spot small objects, we found it crucial to include a ``magnifying glass'' feature that doubles the resolution of a worker's currently selected region.

\myparagraph{Instance Segmentation} Fig.~\ref{fig:ui_pipeline}(c) shows our user interface for instance segmentation. We modified source code from the OpenSurfaces project \cite{bell13opensurfaces}, which defines a single AMT task for segmenting multiple regions of a homogenous material in real-scenes. In our case, we define a single task for segmenting a single object instance labeled from the previous annotation stage. To aid the segmentation process, we added a visualization of the object category icon to remind workers of the category to be segmented. Crucially, we also added zoom-in functionality to allow for efficient annotation of small objects and curved boundaries. In the previous annotation stage, to ensure high coverage of all object instances, we used multiple workers to label all instances per image. We would like to segment {\em all} such object instances, but instance annotations across different workers may refer to different or redundant instances. To resolve this correspondence ambiguity, we sequentially post AMT segmentation tasks, ignoring instance annotations that are already covered by an existing segmentation mask.

\begin{figure*}\centering
  \includegraphics[width=\textwidth]{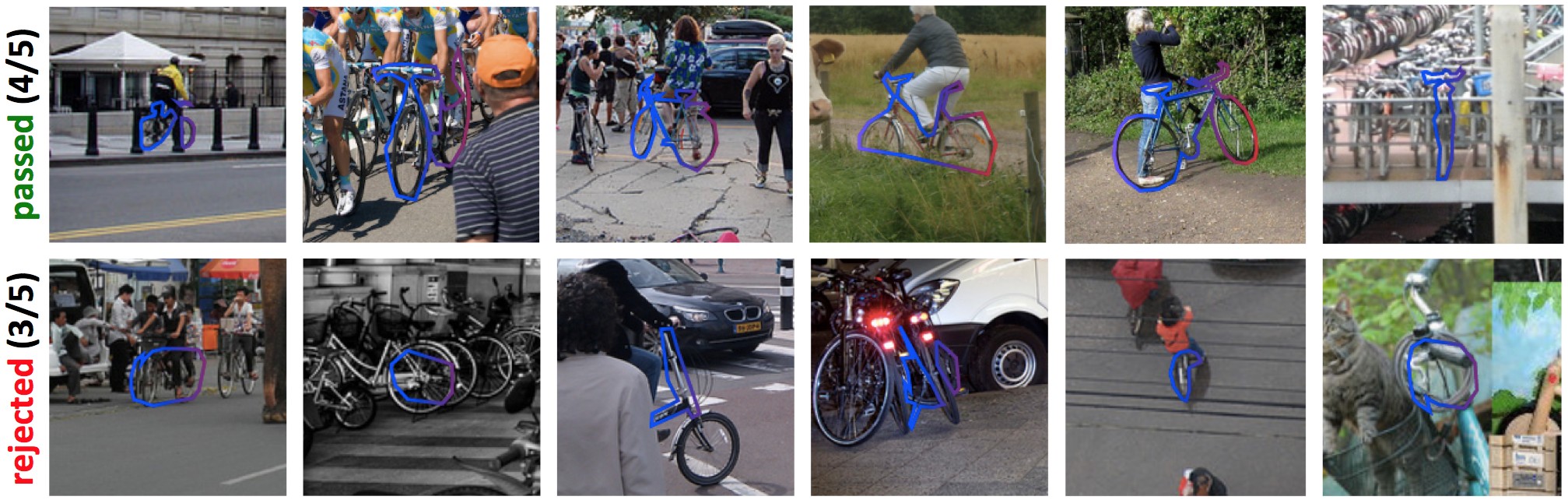}
  \caption{Examples of borderline segmentations that passed (top) or were rejected (bottom) in the verification stage.\label{verification}}
\end{figure*}

\myparagraph{Segmentation Verification} Fig.~\ref{fig:ui_pipeline}(d) shows our user interface for segmentation verification. Due to the time consuming nature of the previous task, each object instance is segmented only once. The purpose of the verification stage is therefore to ensure that each segmented instance from the previous stage is of sufficiently high quality. Workers are shown a grid of 64 segmentations and asked to select poor quality segmentations. Four of the 64 segmentation are known to be bad; a worker must identify 3 of the 4 known bad segmentations to complete the task. Each segmentation is initially shown to 3 annotators. If any of the annotators indicates the segmentation is bad, it is shown to 2 additional workers. At this point, any segmentation that doesn't receive at least 4 of 5 favorable votes is discarded and the corresponding instance added back to the pool of unsegmented objects. Examples of borderline cases that either passed (4/5 votes) or were rejected (3/5 votes) are shown in Fig.~\ref{verification}.

\myparagraph{Crowd Labeling} Fig.~\ref{fig:ui_pipeline}(e) shows our user interface for crowd labeling. As discussed, for images containing ten object instances or fewer of a given category, every object instance was individually segmented. In some images, however, the number of instances of a given category is much higher. In such cases crowd labeling provided a more efficient method for annotation. Rather than requiring workers to draw exact polygonal masks around each object instance, we allow workers to ``paint'' all pixels belonging to the category in question. Crowd labeling is similar to semantic segmentation as object instance are not individually identified. We emphasize that crowd labeling is only necessary for images containing more than ten object instances of a given category.

\section*{Appendix II: Object \& Scene Categories}

Our dataset contains 91 object categories (the 2014 release contains segmentation masks for 80 of these categories). We began with a list of frequent object categories taken from WordNet, LabelMe, SUN and other sources as well as categories derived from a free recall experiment with young children. The authors then voted on the resulting 272 categories with the aim of sampling a diverse and computationally challenging set of categories; see \S\ref{sec:image_collection} for details. The list in Table \ref{tbl:category_list} enumerates those 272 categories in descending order of votes. As discussed, the final selection of 91 categories attempts to pick categories with high votes, while keeping the number of categories per super-category (animals, vehicles, furniture, etc.) balanced.

As discussed in \S\ref{sec:image_collection}, in addition to using object-object queries to gather non-iconic images, object-scene queries also proved effective. For this task we selected a subset of 40 scene categories from the SUN dataset that frequently co-occurred with object categories of interest. Table \ref{tbl:scene_category_list} enumerates the 40 scene categories (evenly split between indoor and outdoor scenes).


\newpage
\bibliographystyle{IEEEtran}
\bibliography{coco}


\newpage


\begin{figure*}\centering
    \begin{subfigure}[b]{0.48\textwidth}
    \includegraphics[width=\textwidth]{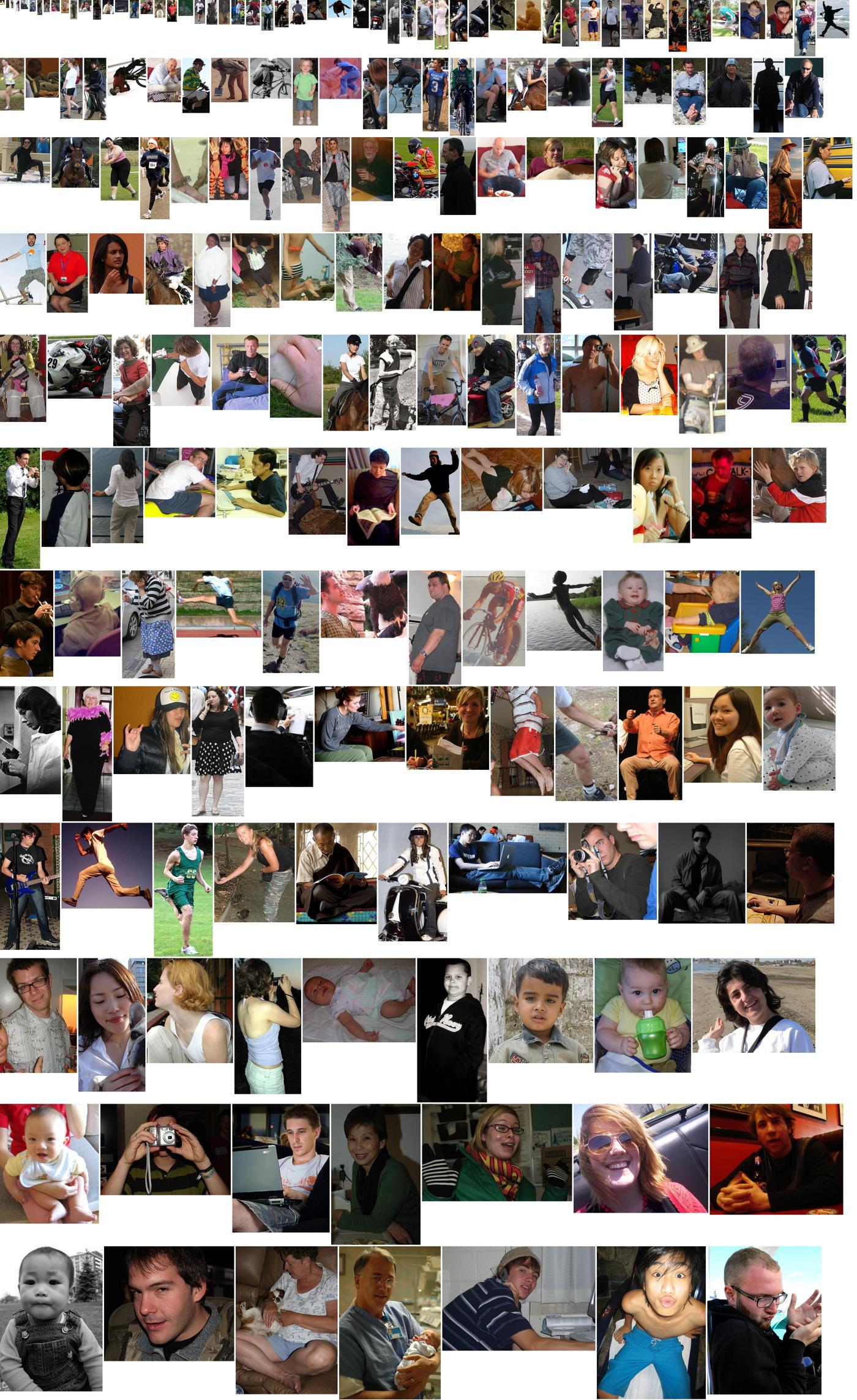}
    \label{fig:train_more_data_pascal_person}\caption{PASCAL VOC.}
   \end{subfigure}\quad
   \begin{subfigure}[b]{0.48\textwidth}
    \includegraphics[width=\textwidth]{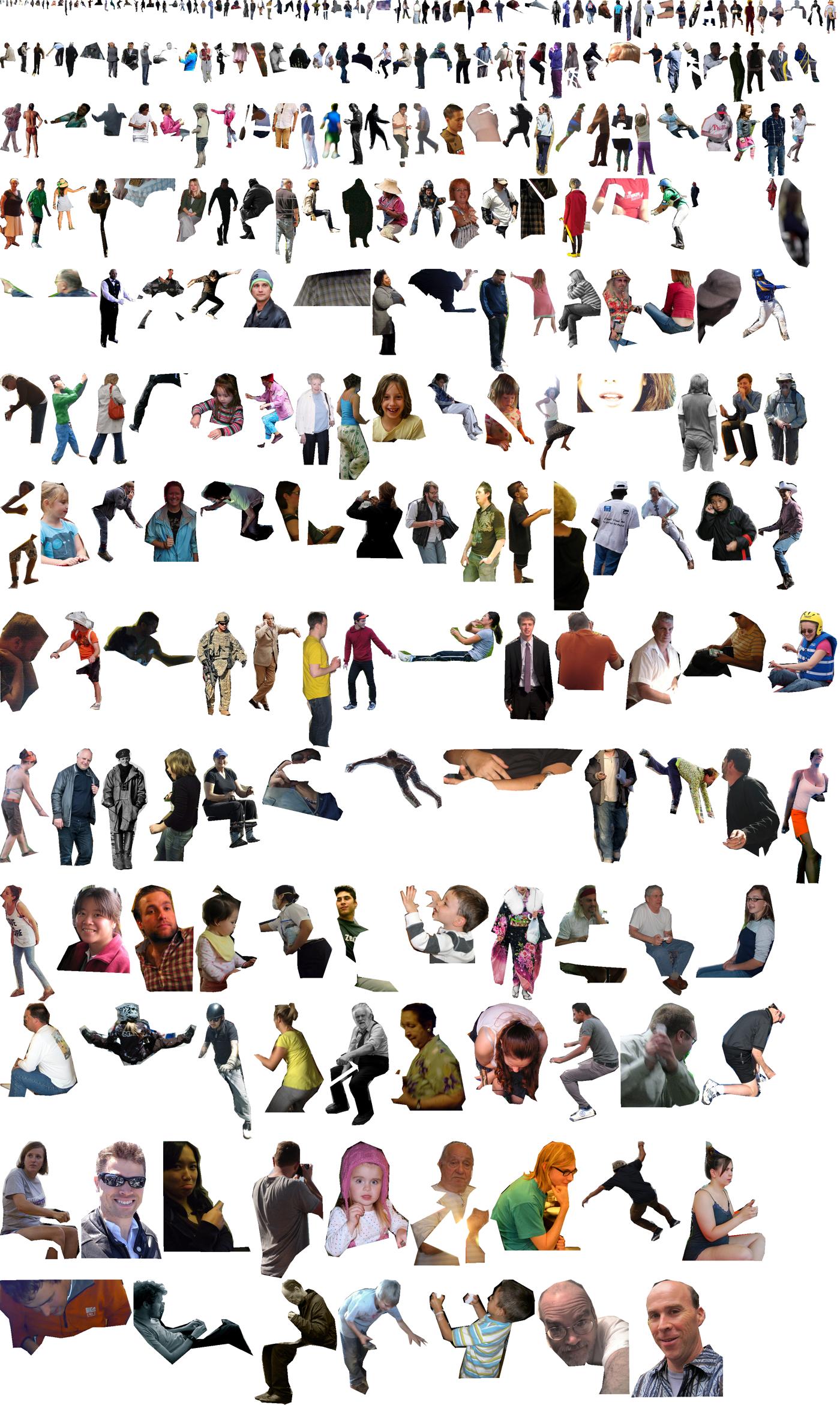}
    \label{fig:train_more_data_coco_person}\caption{\COCO.}
   \end{subfigure}
\caption{Random person instances from PASCAL VOC and \COCO. At most one instance is sampled per image.\label{fig:visualization_person}}\vspace{2mm}
\end{figure*}

\begin{table*}
\resizebox{\textwidth}{!}{
\begin{tabular}{c c c c c c c c c c }
\bf person & \bf bicycle & \bf car & \bf motorcycle & \bf bird & \bf cat & \bf dog & \bf horse & \bf sheep & \bf bottle \\
\bf chair & \bf couch & \bf potted plant & \bf tv & \bf cow & \bf airplane & \bf hat$^*$ & license plate & \bf bed & \bf laptop \\
fridge & \bf microwave & \bf sink & \bf oven & \bf toaster & \bf bus & \bf train & \bf mirror$^*$ & \bf dining table & \bf elephant \\
\bf banana & bread & \bf toilet & \bf book & \bf boat & \bf plate$^*$ & \bf cell phone & \bf mouse & \bf remote & \bf clock \\
face & hand & \bf apple & \bf keyboard & \bf backpack & steering wheel & \bf wine glass & chicken & \bf zebra & \bf shoe$^*$ \\
eye & mouth & \bf scissors & \bf truck & \bf traffic light & \bf eyeglasses$^*$ & \bf cup & \bf blender$^*$ & \bf hair drier & wheel \\
\bf street sign$^*$ & \bf umbrella & \bf door$^*$ & \bf fire hydrant & \bf bowl & teapot & \bf fork & \bf knife & \bf spoon & \bf bear \\
headlights & \bf window$^*$ & \bf desk$^*$ & computer & \bf refrigerator & \bf pizza & squirrel & duck & \bf frisbee & guitar \\
nose & \bf teddy bear & \bf tie & \bf stop sign & \bf surfboard & \bf sandwich & pen/pencil & \bf kite & \bf orange & \bf toothbrush \\
printer & pans & head & \bf sports ball & \bf broccoli & \bf suitcase & \bf carrot & chandelier & \bf parking meter & fish \\
\bf handbag & \bf hot dog & stapler & basketball hoop & \bf donut & \bf vase & \bf baseball bat & \bf baseball glove & \bf giraffe & jacket \\
\bf skis & \bf snowboard & table lamp & egg & door handle & power outlet & hair & tiger & table & coffee table \\
\bf skateboard & helicopter & tomato &tree & bunny & pillow & \bf tennis racket& \bf cake & feet & \bf bench \\
chopping board & washer & lion & monkey & \bf hair brush$^*$ & light switch & arms & legs & house & cheese \\
goat & magazine & key & picture frame & cupcake & fan (ceil/floor) & frogs & rabbit & owl & scarf \\
ears & home phone & pig & strawberries & pumpkin & van & kangaroo & rhinoceros & sailboat & deer \\
playing cards & towel & hyppo & can & dollar bill & doll & soup & meat & window & muffins \\
tire & necklace & tablet & corn & ladder & pineapple & candle & desktop & carpet & cookie \\
toy cars & bracelet & bat & balloon & gloves & milk & pants & wheelchair & building & bacon \\
box & platypus & pancake & cabinet & whale & dryer & torso & lizard & shirt & shorts \\
pasta & grapes & shark & swan & fingers & towel & side table & gate & beans & flip flops \\
moon & road/street & fountain & fax machine & bat & hot air balloon & cereal & seahorse & rocket & cabinets \\
basketball & telephone  & movie (disc) & football & goose & long sleeve shirt & short sleeve shirt & raft & rooster & copier \\
radio & fences & goal net & toys & engine & soccer ball & field goal posts & socks & tennis net & seats \\
elbows & aardvark & dinosaur & unicycle & honey & legos & fly & roof & baseball & mat \\
ipad & iphone & hoop & hen & back & table cloth & soccer nets & turkey & pajamas & underpants \\
goldfish & robot & crusher & animal crackers & basketball court & horn & firefly & armpits & nectar & super hero costume \\
jetpack & robots & & & & & & & &
\end{tabular} }
\caption{Candidate category list (272). {\bf Bold}: selected categories (91). {\bf Bold$^*$}: omitted categories in 2014 release (11).\label{tbl:category_list}}
\end{table*}

\begin{figure*}\centering
   \begin{subfigure}[b]{0.48\textwidth}
    \includegraphics[width=\textwidth]{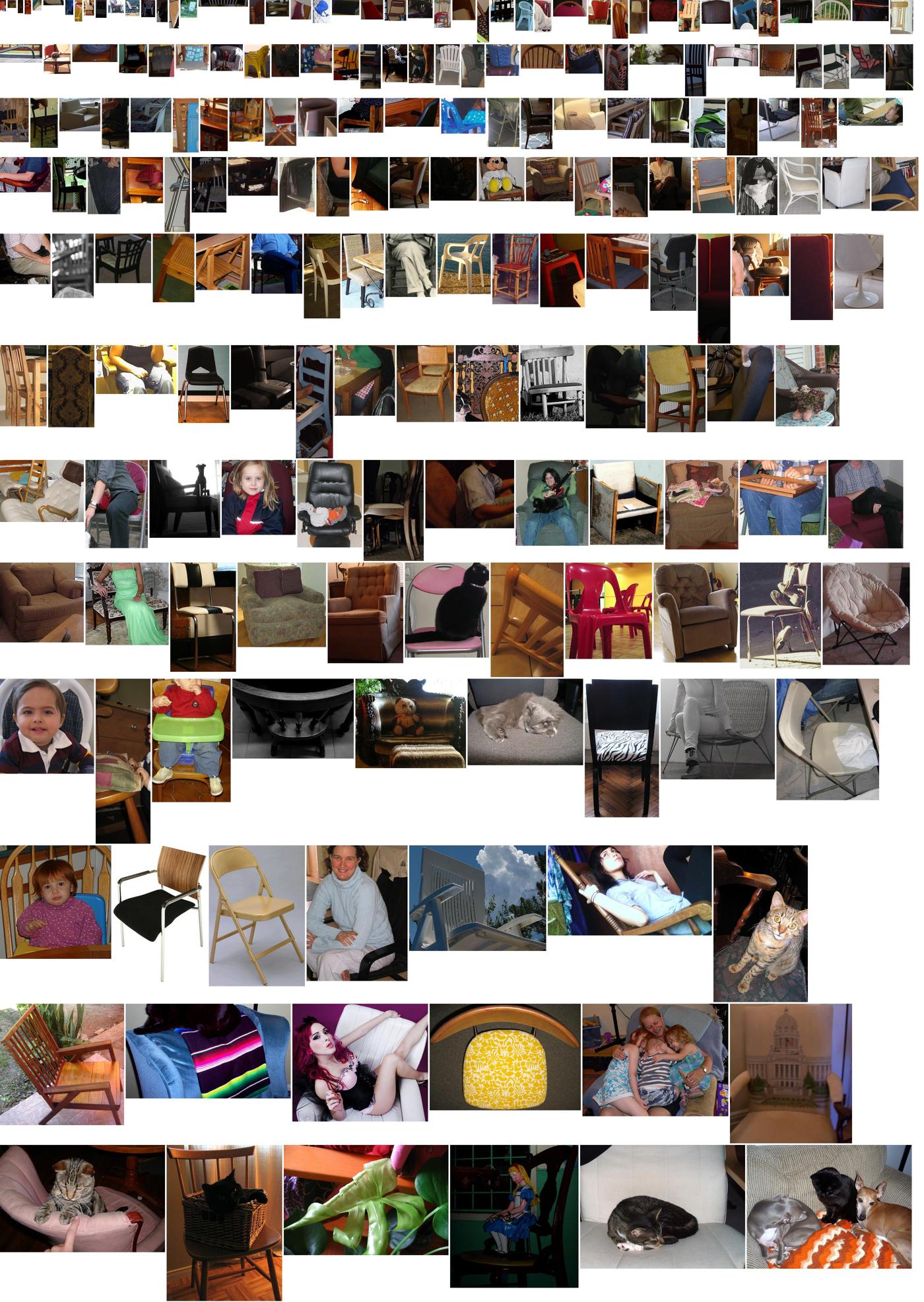}
    \label{fig:train_more_data_pascal_chair}\vspace{-.5cm}
    \caption{PASCAL VOC.}
   \end{subfigure}  \quad
   \begin{subfigure}[b]{0.48\textwidth}
    \includegraphics[width=\textwidth]{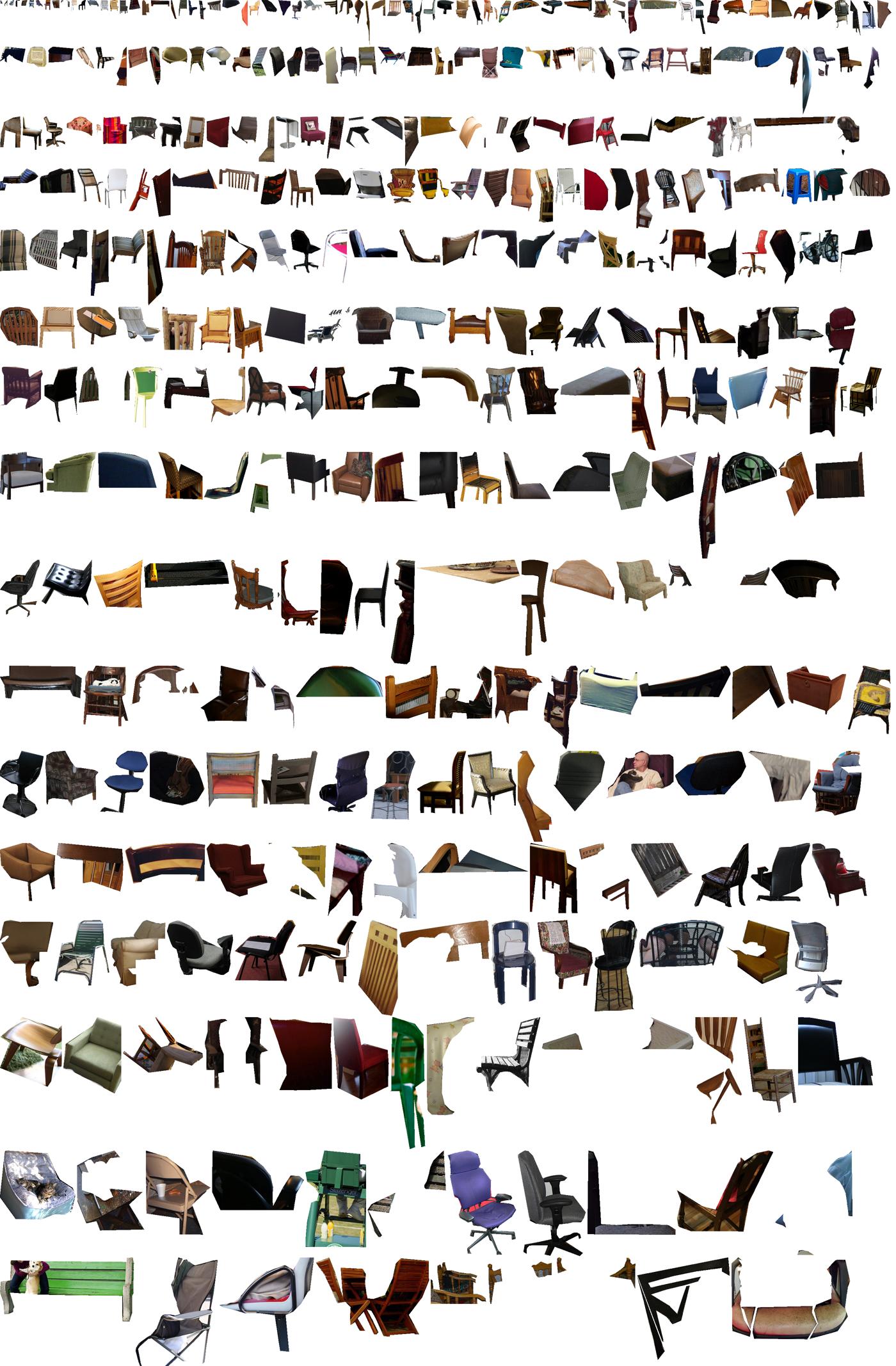}
    \label{fig:train_more_data_coco_chair}\vspace{-.5cm}
    \caption{\COCO.}
   \end{subfigure}
  \caption{Random chair instances from PASCAL VOC and \COCO. At most one instance is sampled per image.\label{fig:visualization_chair}}
\end{figure*}

\begin{table*}\centering
\resizebox{.95\textwidth}{!}{
\begin{tabular}{c c c c c c c c c c }
library & church & office & restaurant & kitchen & living room & bathroom & factory & campus & bedroom \\
child's room & dining room & auditorium & shop & home & hotel & classroom & cafeteria & hospital room & food court \\
street & park & beach & river & village & valley & market & harbor & yard & parking lot \\
lighthouse & railway & playground & swimming pool & forest & gas station & garden & farm & mountain & plaza
\end{tabular} }
\caption{Scene category list.\label{tbl:scene_category_list}}
\end{table*}


\end{document}